\documentclass[lettersize,journal]{IEEEtran}
\usepackage{amsmath,amsfonts}
\usepackage{array}
\usepackage[caption=false,font=normalsize,labelfont=sf,textfont=sf]{subfig}
\usepackage{textcomp}
\usepackage{stfloats}
\usepackage{url}
\usepackage{verbatim}
\usepackage{graphicx}

\usepackage{cite}

\usepackage{multirow}%
\usepackage{amsmath,amssymb,amsfonts}%
\usepackage{amsthm}%
\usepackage{mathrsfs}%
\usepackage{xcolor}%
\usepackage{textcomp}%
\usepackage{manyfoot}%
\usepackage{booktabs}%
\usepackage{algorithm}%
\usepackage{algorithmicx}%
\usepackage{algpseudocode}%
\usepackage{listings}%
\usepackage{anyfontsize}
\usepackage{color}
\usepackage{bbding}
\usepackage{longtable}
\usepackage{supertabular}
\usepackage{colortbl}
\usepackage{hyperref}

\newtheorem{theorem}{Theorem}%

\newtheorem{proposition}[theorem]{Proposition}%

\begin{document}

\title{See Further for Parameter Efficient Fine-tuning by Standing on the Shoulders of Decomposition}

\author{Chongjie Si, Xiaokang Yang,~\IEEEmembership{~Fellow,~IEEE}, Wei Shen
\thanks{C. Si, X. Yang, and W. Shen are with
MoE Key Lab of Artificial Intelligence, AI Institute, Shanghai Jiao Tong University, Shanghai, China.}
\thanks{Email: \{chongjiesi, xkyang, wei.shen\}@sjtu.edu.cn}
\thanks{Codes are available at \url{https://github.com/Chongjie-Si/Subspace-Tuning}.}
}

\markboth{Journal of \LaTeX\ Class Files,~Vol.~14, No.~8, August~2021}%
{Shell \MakeLowercase{\textit{et al.}}: A Sample Article Using IEEEtran.cls for IEEE Journals}


\maketitle

\begin{abstract}

The rapid expansion of large foundation models within the pre-training and fine-tuning framework has underscored that larger models often yield better results. 
However, the scaling up of large foundation models has led to soaring costs in fine-tuning and parameter storage, rendering extensive adaptations impractical. This challenge has sparked the development of parameter-efficient fine-tuning (PEFT), which focuses on optimizing a select subset of parameters while keeping the rest fixed, significantly lowering computational and storage overheads. While recent years have witnessed a significant success in PEFT, a deep understanding of the fundamental principles behind these methods remains unexplored. 
To this end, here we take the first step to unify all approaches by dissecting them from a decomposition perspective. We initiate a comprehensive mathematical analysis of these methods, allowing us to delve deeply into their underlying mechanisms, and we explore the reasons behind the variations in performance among different techniques.
Furthermore, inspired by our theoretical analysis, we introduce two novel PEFT methods alongside a simple yet effective framework designed to enhance the performance of PEFT techniques across various applications. 
Our empirical validations, conducted across multiple datasets, demonstrate the efficacy of these methods, showcasing both theoretical validity and practical performance improvements under the guidance of our analytical findings. We believe our work will deepen researchers' understanding of PEFT and other techniques, prompting further contemplation and advancing the research across the whole community.

\end{abstract}


\section{Introduction}

The emergence of foundation models, as referenced in multiple studies \cite{brown2020language, radford2019language, radford2021learning, devlin2018bert, liu2019roberta}, has fundamentally altered the landscape of artificial intelligence, demonstrating substantial effectiveness across a variety of domains. 
For instance, Segment Anything Model (SAM) \cite{kirillov2023segment} has been widely implemented across a variety of visual tasks \cite{si2024tendency,zhang2023customized,zhang2023comprehensive}, and Generative Pre-trained Transformer (GPT) \cite{brown2020language,radford2019language} has even seamlessly integrated into our daily lives, evolving into an exceedingly practical tool \cite{achiam2023gpt,waisberg2023gpt,mao2023gpteval}.
Traditionally, the adaptation of pre-trained models to specific downstream tasks required fully fine-tuning of all parameters \cite{ma2024segment, raffel2020exploring, qiu2020pre}. 
However, as the complexity and size of these models have increased, this traditional approach to fine-tuning has become less feasible, both from a computational and resource standpoint.

In response to these challenges, there has been a pivot towards developing more efficient techniques \cite{chen2024parameter, guo2020parameter, he2021towards, hu2021lora}, collectively known as parameter-efficient fine-tuning (PEFT). 
The goal of PEFT is to achieve comparable or even superior performance on downstream tasks by tuning a minimal number of parameters compared with fully fine-tuning.
Presently, PEFT strategies can be categorized into three predominant groups \cite{liu2024dora, ding2023parameter}, each with its distinctive mechanisms and intended use cases.

Firstly, adapter-based methods, as discussed in several works \cite{houlsby2019parameter, chen2022adaptformer, luo2023towards, he2021towards, mahabadi2021parameter, karimi2021compacter}, involve the insertion of small, trainable linear modules within the pre-existing network architectures.
These modules are designed to adapt the model's outputs without changing the original network weights. 
Secondly, the prompt-based approaches \cite{lester2021power, razdaibiedina2023residual, wang2023non, shi2023dept, fischer2024prompt} make use of mutable soft tokens placed at the beginning of inputs.
This strategy focuses on fine-tuning these prompts to steer the model's behavior during specific tasks.
Thirdly, low-rank adaptation approaches like LoRA derivatives \cite{hu2021lora, liu2024dora, hyeon2021fedpara, qiu2023controlling, renduchintala2023tied, kopiczko2023vera, yeh2023navigating, zhang2022adaptive, si2024flora, si2024unleashing} are applied to network weights during fine-tuning, enhancing their adaptability while maintaining overall compatibility with the pre-trained settings.
Additionally, the landscape of PEFT is enriched by other innovative methods such as BitFit \cite{zaken2021bitfit, lawton2023neural}, which focus solely on fine-tuning the bias terms. Collectively, these diverse strategies significantly augment the adaptability and efficiency of models, enabling them to meet specific task requirements without the need for extensive retraining. 
Through these developments, the whole community continues to evolve towards more sustainable and manageable model training methodologies.

However, despite that recent years have witnessed significant advancements in PEFT \cite{han2024parameter,he2021towards, fu2023effectiveness}, the mathematical foundations underpinning different PEFT methods have scarcely been studied. 
Moreover, the performance differences between various PEFT methods and the reasons behind these differences have not been systematically explored. 
This lack of theoretical depth limits our understanding of the potential advantages and limitations of these methods, hindering their optimization and innovation in practical applications. 
Therefore, conducting theoretical research in this field will be crucial for advancing PEFT technologies, providing a fundamental basis for selecting and designing more efficient strategies.

In this paper, we undertake a pioneering theoretical examination of PEFT techniques, leveraging insights from decomposition theory including matrix (decomposition) and subspace (decomposition) theory. 
We introduce a novel framework termed \textit{subspace tuning}, which encapsulates all known PEFT methods under a unified theory.
The subspace tuning method primarily focuses on adjusting the subspace of the original parameter, involving both the reconstruction and the extension of subspaces. 
We delve into how different methods manipulate subspaces and elucidate the mathematical principles underlying each approach from the perspective of decomposition theory. 
Additionally, we analyze why these methods result in performance differences, providing a comprehensive theoretical foundation to understand the dynamics within different PEFT strategies.

Furthermore, inspired by our theoretical analysis, we propose two novel PEFT methods. 
Compared to existing techniques, these new approaches achieve performance close to fully fine-tuning with only 0.02\% parameters. 
Additionally, we introduce an effective framework that enhances the performance of methods such as LoRA without introducing additional training parameters. 
This framework provides a practical solution to optimize PEFT methodologies, thereby extending their applicability and effectiveness in resource-constrained environments. 
Extensive experiments are conducted to validate our theoretical propositions by testing more than ten methods on three different models.
They not only confirm the robustness of our theoretical insights but also demonstrate the efficacy of the methods and framework we proposed.

We hope that our research could significantly inspire further studies in PEFT and other related communities \cite{mudrakarta2018k,si2024partial,jovanovic2024trends,feng2023peft,si2023multi,mahabadi2021parameter}, catalyzing advancements and influencing developments across the broader artificial intelligence landscape.

\section{Subspace Tuning}\label{sec result overview}

\begin{figure*}[!ht]
\centering
    \includegraphics[width=0.8\textwidth]{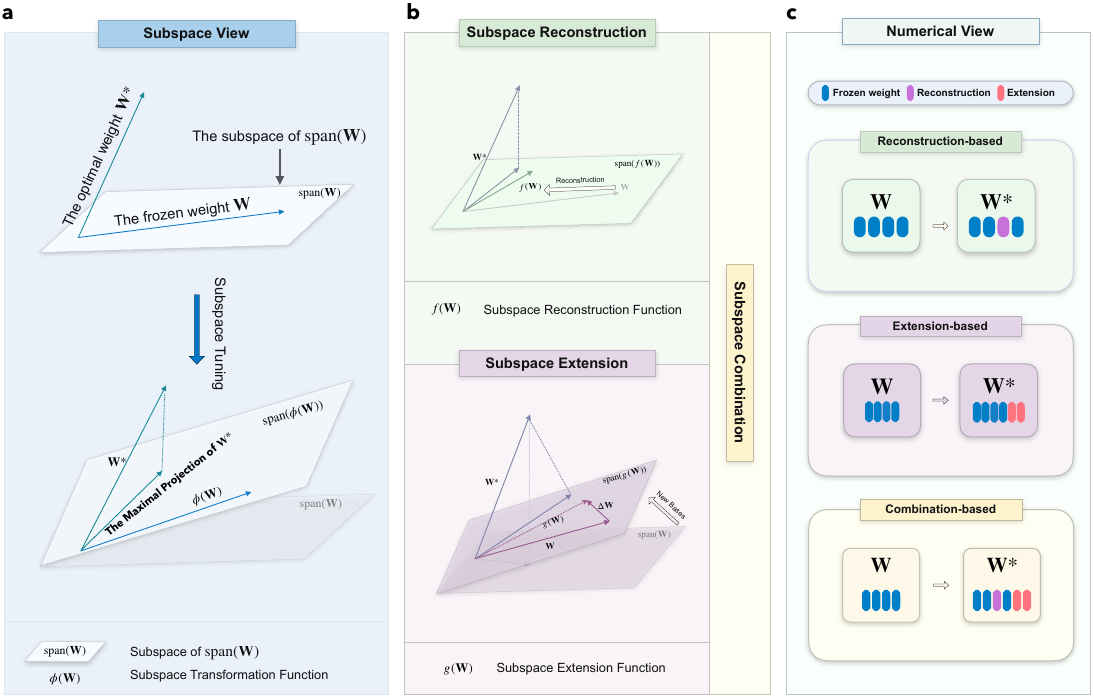}
    \caption{Framework of subspace tuning. \textbf{a}, Subspace tuning endeavors to identify the maximal projection of the optimal weight $\mathbf{W}^{*}$ onto the subspace spanned by the bases of $\phi(\mathbf{W})$. Here, $\phi(\mathbf{W})$ denotes the subspace transformation of the original frozen weight $\mathbf{W}$. \textbf{b}, Subspace reconstruction involves rescaling the subspace of $\mathbf{W}$ to approximate $\mathbf{W}^{*}$, or to construct a new subspace derived from the original. Subspace extension seeks to adjust the subspace of the original weight $\mathbf{W}$ such that it approaches or even encompasses $\mathbf{W}^{*}$. Subspace combination encompasses both the reconstruction and extension of subspaces. \textbf{c}, A numerical perspective on subspace tuning. Reconstruction involves modifying the frozen parameters, while extension entails adding new tunable parameters.}
    \label{fig:framework}
\end{figure*}

\makeatletter
\newcommand{\subfig}[2]{\hyperref[#1]{\ref*{#1}#2}}
\makeatother

Consider ${\mathbf{W}}\in\mathbb{R}^{n\times m}$ as the frozen weight matrix of a layer in a pre-trained neural network, with $n\leq m$ without loss of generality.
The performance of the model parameterized by $\mathbf{W}$ on a specific task is quantified by a performance function $\mathcal{P}:\mathbb{R}^{n\times m}\rightarrow\mathbb{R}$, where a higher value indicates better performance.
Assume there exists an optimal weight matrix $\mathbf{W}^{*}\in\mathbb{R}^{n\times m}$ for the task at hand \cite{ding2023parameter}, satisfying
\begin{equation}
\mathcal{P}(\mathbf{W}^{*}) = \max_{\bar{\mathbf{W}}\in\mathbb{R}^{n\times m}} \mathcal{P}(\bar{\mathbf{W}}).
\end{equation}
Typically, $\mathbf{W}^{*}$ can be thought as the weights by fully fine-tuning of the pre-trained model \cite{wang2024loraga,si2024unleashing}.

The objective of PEFT methods is to approximate $\mathbf{W}^{*}$ while training only a small subset of parameters \cite{meng2024pissa,wang2024loraga,si2024unleashing,wang2024lorapro,hao2024flora}. 
This objective can be formalized as finding a transformation function $\phi:\mathbb{R}^{n\times m}\rightarrow\mathbb{R}^{n\times m}$ such that
\begin{equation}
\min_{\phi} \ \ell\left(\mathbf{W}^{*}, \phi(\mathbf{W})\right),
\label{eq loss W and W*}
\end{equation}
where $\ell:\mathbb{R}^{n\times m} \times \mathbb{R}^{n\times m} \rightarrow \mathbb{R}$ is a loss function measuring the discrepancy between $\mathbf{W}^{}$ and $\phi(\mathbf{W})$, such as the Frobenius norm:
\begin{equation}
\ell\left(\mathbf{W}^{*}, \phi(\mathbf{W})\right) = \|\mathbf{W}^{*} - \phi(\mathbf{W})\|_F^2.
\end{equation}

In previous works, the function $\phi$ has been conceptualized as modifications to each element of the matrix $\mathbf{W}$ \cite{ding2023parameter}. 
While this characterization is accurate, it is overly general and does not adequately capture the underlying logic of each approach.
Indeed, from the perspective of matrix theory, adjusting $\mathbf{W}$ involves modifying its associated subspaces.
Therefore, we interpret all PEFT methods as forms of \textbf{Subspace Tuning} (Fig. \subfig{fig:framework}{a}).
Specifically, we consider $\phi$ as a function that transforms the subspaces associated with $\mathbf{W}$, and Eq. (\ref{eq loss W and W*}) then aims to find $\phi$ such that $\phi(\mathbf{W})$ approximates $\mathbf{W}^{*}$ by adjusting or extending the subspaces of $\mathbf{W}$. There are two primary strategies to achieve this:

\begin{itemize}
    \item \textbf{Subspace Reconstruction}: Modify the subspaces associated with $\mathbf{W}$ through a transformation $f: \mathbb{R}^{n\times m} \rightarrow \mathbb{R}^{n\times m}$ to better align with the subspaces of $\mathbf{W}^{*}$. The transformation function is then $\phi(\mathbf{W}) = f(\mathbf{W})$.
    
    \item \textbf{Subspace Extension}: Introduce additional subspaces by combining $\mathbf{W}$ with an additive component, typically represented as $g: \mathbb{R}^{n\times m} \rightarrow \mathbb{R}^{n\times m}$, resulting in $\phi(\mathbf{W}) = g(\mathbf{W})$.
\end{itemize}
These strategies can be mathematically unified as
\begin{equation}
\phi(\mathbf{W}) = g(f(\mathbf{W})),
\end{equation}
where $f$ represents subspace reconstruction and $g$ represents subspace extension (Fig. \subfig{fig:framework}{b}). Based on these strategies, we classify existing PEFT methods into three categories:

\begin{itemize}
\item \textbf{Reconstruction-based Methods}: Adjust the subspaces of $\mathbf{W}$ through transformations, aiming to align them with those of $\mathbf{W}^{*}$.
\item \textbf{Extension-based Methods}: Augment $\mathbf{W}$ by introducing additional components $\Delta \mathbf{W}$, expanding the subspace to include new directions.
\item \textbf{Combination-based Methods}: Integrate both reconstruction and extension strategies to adjust and expand the subspaces simultaneously.
\end{itemize}

We will briefly introduce each category and explore the underlying mathematical principles of corresponding methods. The full details are left to Methods.
To simplify the notations for the following sections, let $\mathbf{A}\in\mathbb{R}^{n\times r}$ and $\mathbf{B}\in\mathbb{R}^{r\times m}$ ($r\ll n,m$) be two matrices that map the subspace to different dimensions, with the rank being $r$. $\mathbf{D}\in\mathbb{R}^{n\times m}$ represents a (rectangular) diagonal matrix. 
For a specific matrix $\mathbf{W}_0\in\mathbb{R}^{n\times m}$, we use $\mathbf{W}_0^{\dagger}$ to represent its Moore-Penrose pseudo-inverse, and $\mathbf{U}_0\in\mathbb{R}^{n\times n}$ and $\mathbf{V}_0\in\mathbb{R}^{m\times m}$ to represent its left and right singular vectors from its singular vector decomposition, with $\mathbf{\Sigma}_0\in\mathbb{R}^{n\times m}$ being the corresponding singular values. 
All the notations are included in Table \ref{tab:notations}.

\begin{table}[ht]
    \centering
    \caption{Summary of major notations.}
    \begin{tabular}{c c}
    \toprule
        \textbf{Notation} & \textbf{Mathematical Meanings} \\\hline
       $\mathcal{P}(\cdot)$  & The Performance of a Model \\\hline
       
       $\mathbf{W} \in \mathbb{R}^{n\times m}$ & Frozen Weight Matrix \\\hline
       
       $\phi(\cdot)$  & Subspace Transformation Function \\
       
       $f(\cdot)$ & Subspace Reconstruction Function\\
    
       $g(\cdot)$ & Subspace Extension Function \\ \hline
       $\mathbf{A}\in\mathbb{R}^{n\times r}$ & Down Projection Matrix \\ 
       $\mathbf{B}\in\mathbb{R}^{r\times m}$ & Up Projection Matrix \\
       $\mathbf{D}\in\mathbb{R}^{n\times m}$ & (Rectangle) Diagonal Matrix \\\hline
       
       \multirow{2}{*}{$\mathbf{W}^{\dagger}\in\mathbb{R}^{m\times n}$} & Moore-Penrose Pseudo-inverse \\
       &  of the Matrix $\mathbf{W} \in \mathbb{R}^{n\times m}$\\\hline
       $\mathbf{U}\in\mathbb{R}^{n\times n}$ & Left Singular Vectors\\
       $\mathbf{V}\in\mathbb{R}^{m\times m}$ & Right Singular Vectors\\
       $\mathbf{\Sigma}\in\mathbb{R}^{n\times m}$ & Singular Values\\
         \bottomrule
    \end{tabular}
    \label{tab:notations}
\end{table}

\section{Subspace Reconstruction}

In this section, we focus on methods that reconstruct the subspaces associated with the weight matrix $\mathbf{W}$. These methods aim to modify $\mathbf{W}$ through the subspace transformation, specifically expressed as $\phi(\mathbf{W}) = f(\mathbf{W})$, to better approximate the optimal weight matrix $\mathbf{W}^{*}$.

\begin{figure*}[!ht]
    \centering
    \includegraphics[width=\textwidth]{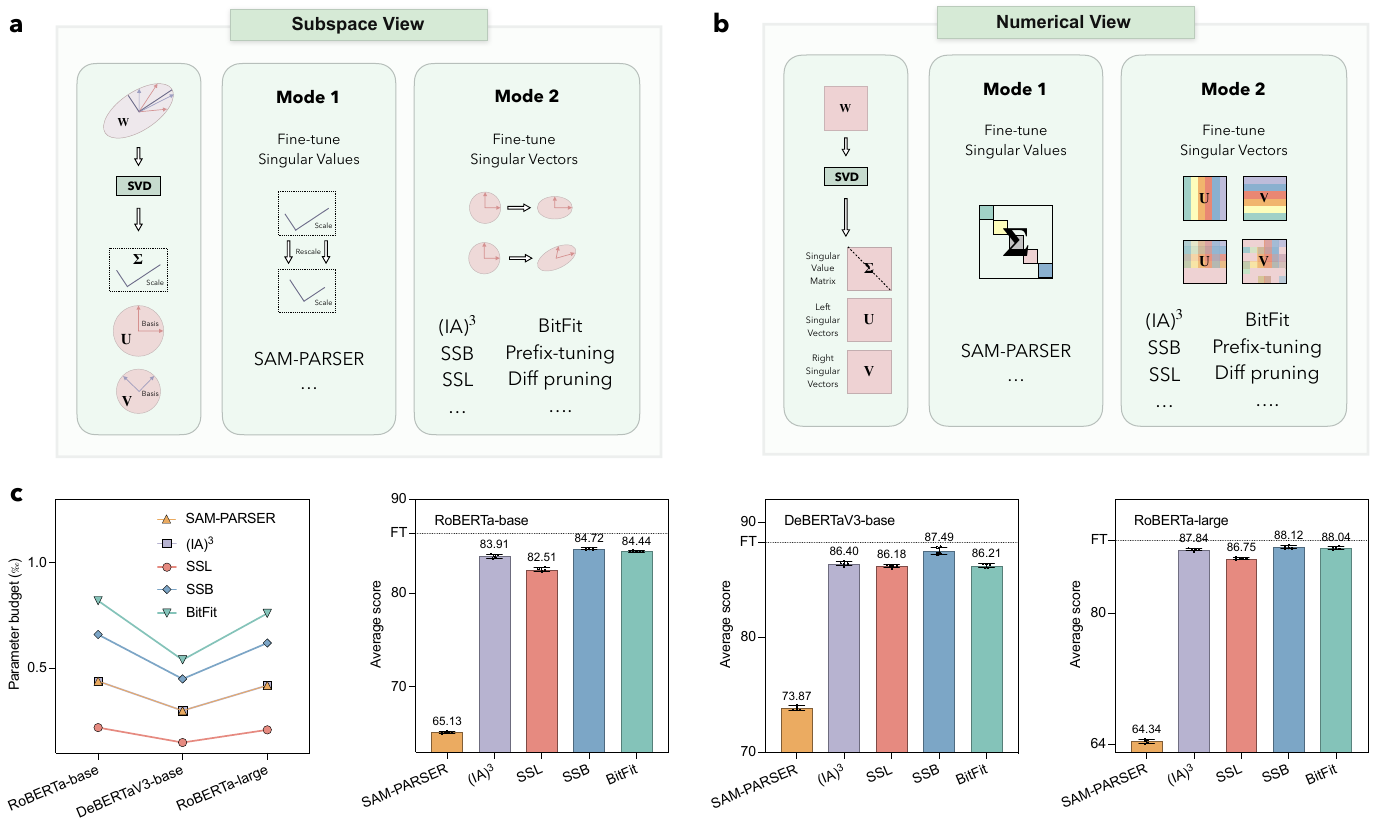}
    \caption{\textbf{a}, Subspace view of reconstruction-based methods. Fine-tuning the singular values involves rescaling the weights, while fine-tuning the singular vectors effectively reconstructs the subspace. \textbf{b}. Numerical view of reconstruction-based methods. We correspond adjustments in the subspace directly to their numerical adjustments. \textbf{c}, The performance of reconstruction-based methods. With less than 0.1\% of the parameters of the pretrained model, SSL and SSB can achieve up to 99\% of the performance of fully fine-tuning. The horizontal dashed line parallel to the x-axis, labeled FT, represents the performance of fully fine-tuning. The average scores of each method are evaluated with three large pretrained models, RoBERTa-base \cite{liu2019roberta}, DeBERTaV3-base \cite{he2021debertav3}, and RoBERTa-large \cite{liu2019roberta} on the GLUE benchmark. Error bars represent the standard error of the mean across five runs.
    }
    \label{fig: reconstruction result}
\end{figure*}

We begin by considering the Singular Value Decomposition (SVD) of $\mathbf{W}$. The SVD is a fundamental technique in matrix theory that decomposes a matrix into its constituent subspaces. Formally, for $\mathbf{W} \in \mathbb{R}^{n \times m}$, its SVD is given by:
\begin{equation}
\mathbf{W} = \mathbf{U} \mathbf{\Sigma} \mathbf{V}^{\mathsf{T}},
\end{equation}
where $\mathbf{U} \in \mathbb{R}^{n \times n}$ is an orthogonal matrix whose columns are the left singular vectors of $\mathbf{W}$, forming an orthonormal basis for the column space of $\mathbf{W}$, $\mathbf{\Sigma} \in \mathbb{R}^{n \times m}$ is a diagonal matrix with non-negative singular values $\sigma_i$ on the diagonal, representing the scaling factors along each principal direction, and $\mathbf{V} \in \mathbb{R}^{m \times m}$ is an orthogonal matrix whose columns are the right singular vectors of $\mathbf{W}$, forming an orthonormal basis for the row space of $\mathbf{W}$.
The goal of subspace reconstruction methods is to adjust the subspaces related to $\mathbf{U}$, $\mathbf{\Sigma}$, or $\mathbf{V}$ to construct a new weight matrix $\phi(\mathbf{W})$ that better approximates $\mathbf{W}^{*}$. 
We categorize these adjustments into two distinct modes (Fig. \subfig{fig: reconstruction result}{a} and \subfig{fig: reconstruction result}{b}).

\begin{itemize}
    \item \textbf{Mode 1, Singular Value Adjustment:}
    This mode entails the modification of the singular values in $\mathbf{\Sigma}$, thereby adjusting the scaling within the respective principal subspaces. Altering these values modifies the significance attributed to each principal component, without affecting the directional properties of the subspaces defined by $\mathbf{U}$ and $\mathbf{V}$.
    \item \textbf{Mode 2, Singular Vector Adjustment:}
    This mode involves adjustments to the singular vectors in \(\mathbf{U}\) and \(\mathbf{V}\), including scaling the subspaces they span or more intricate transformations such as reorientation or reshaping of the subspaces.
    It facilitates a comprehensive adjustment of the matrix structure.
\end{itemize}

\subsection{Singular Value Adjustment}

In Mode 1, we assume that the optimal weight matrix $\mathbf{W}^{}$ shares the same singular vectors as $\mathbf{W}$. That is, we posit that the left and right singular vectors remain unchanged, and only the singular values need to be adjusted. Formally, we write:
\begin{equation}
\mathbf{W}^{*} = \mathbf{U} \mathbf{\Sigma}^{*} \mathbf{V}^{\mathsf{T}},
\end{equation}
where $\mathbf{\Sigma}^{*}$ is a diagonal matrix containing the optimal singular values $\sigma_i^{*}$.
The transformation function $\phi$ then becomes:
\begin{equation}
\phi(\mathbf{W}) = \mathbf{U} \mathbf{\Sigma}' \mathbf{V}^{\mathsf{T}},
\end{equation}
where $\mathbf{\Sigma}'$ is a learnable diagonal matrix that we aim to optimize. 
This is what SAM-PARSER \cite{peng2024sam} actually does.

Specifically, SAM-PARSER targets the weight matrix of the SAM architecture’s ``neck'' component, which consists of two layers of dimension $256 \times 256$. 
The reconstruction process is based on optimizing the 512 singular values corresponding to these two layers.
However, this method relies on the assumption that the singular vectors of $\mathbf{W}$ and $\mathbf{W}^{*}$ are identical, which may not hold in practice.
The limitation arises because the adjustment only scales the principal components without changing their directions, potentially restricting the ability to approximate $\mathbf{W}^{*}$ accurately.
Consequently, under equivalent conditions, the performance of the Mode 1 method is anticipated to lag considerably behind that of other techniques. Results in Fig. \subfig{fig: reconstruction result}{c} (Supplementary Tables \ref{tab: roberta base results}-\ref{tab: roberta  large results}) proves that this method is substantially inferior to alternative approaches, even when allowed a larger parameter budget.

\subsection{Singular Value Adjustment}

Mode 2 methods specifically focus on the manipulation of subspaces spanned by the singular vectors of weight matrices.
This can be formalized as follows:
\begin{equation}
\phi(\mathbf{W}) = \mathcal{T}_1(\mathbf{U}) \mathbf{\Sigma} \mathcal{T}_2(\mathbf{V})^{\mathsf{T}}.
\end{equation}
where $\mathcal{T}_1: \mathbb{R}^{n \times n} \rightarrow \mathbb{R}^{n \times n}$ and $\mathcal{T}_2: \mathbb{R}^{m \times m} \rightarrow \mathbb{R}^{m \times m}$, which may be linear or nonlinear functions. 
It allows for scaling, rotations, reflections, and other transformations like element-wise nonlinear mapping of the singular vectors, providing the greatest flexibility in adjusting the subspaces.

\subsubsection{Scaling}

We start from scaling the subspaces.
Let $\mathbf{D}_1 \in \mathbb{R}^{n \times n}$ and $\mathbf{D}_2 \in \mathbb{R}^{m \times m}$ be diagonal scaling matrices.

\noindent
\textbf{Scale the Column Space: }
If we scale the column space of singular vectors by assigning distinct weights to each vector, the reconstructed weight matrix $\hat{\mathbf{W}}$ can then be obtained as:
\begin{equation}
    \hat{\mathbf{W}} = \mathbf{U}\mathbf{D}_1\mathbf{\Sigma}  \mathbf{D}_2\mathbf{V}^\mathsf{T}  = \mathbf{U}\hat{\mathbf{\Sigma}}\mathbf{V}^\mathsf{T},
\end{equation}
where $\hat{\mathbf{\Sigma}}=\mathbf{D}_1\mathbf{\Sigma}  \mathbf{D}_2$.
Therefore, scaling the column space of singular vectors is just an adjustment of the singular values.

\noindent
\textbf{Scale the Row Space:}
We can also apply distinct weights to each row of the singular vectors. Consequently, the reconstructed weight matrix $\hat{\mathbf{W}}$ is articulated as follows:
\begin{equation}
    \hat{\mathbf{W}}  = \mathbf{D}_1\mathbf{U}\mathbf{\Sigma}  \mathbf{V}^\mathsf{T}\mathbf{D}_2  =  \mathbf{D}_1\mathbf{W}\mathbf{D}_2.
\label{eq adjust the row}
\end{equation}
Thus, scaling the row space spanned by the left and right singular vectors essentially corresponds to scaling both the row and column spaces of the original weight matrix.

From this perspective, some methods can yield more in-depth explanations, such as (IA)$^3$ \cite{liu2022few}. 
In the original paper \cite{liu2022few}, this method seeks to directly modify the activations within the model by introducing a learnable vector $\mathbf{l}\in\mathbb{R}^m$ to rescale the original weight matrix. 
The transformation is implemented via the Hadamard product $\odot$, represented as $\mathbf{l}\odot\mathbf{W}$. 
However, it can equivalently be expressed as $\mathbf{W}\mathbf{D}_2$, where $\mathbf{D}_2\in\mathbb{R}^{m\times m}$ is a diagonal matrix. 
Consequently, this approach actually scales the subspace of the right singular vectors, thereby reconstructing the original weight matrix $\mathbf{W}$.

The results in Fig. \subfig{fig: reconstruction result}{c} demonstrate that merely scaling the subspace of the right singular vectors, i.e., (IA)$^3$, can achieve performance comparable to fully fine-tuning. 
This insight naturally gives rise to an additional adjustment method: Scaling the Subspace of the Left singular vectors (SSL).
If the dimensions of the subspaces spanned by both left and right singular vectors are comparable, the performance of SSL and (IA)$^3$ are expected to be similar, since both methods enhance model adaptation by scaling a singular subspace. This is corroborated by the results shown in Fig. \subfig{fig: reconstruction result}{c}.

Further expanding on this concept, we introduce the method of Scaling the Subspace of Both left and right singular vectors (SSB). Theoretically, SSB should outperform both SSL and (IA)$^3$ as it simultaneously scales both subspaces, potentially enhancing the reconstruction quality beyond the capabilities of single-subspace scaling. Results from Fig. \subfig{fig: reconstruction result}{c} and detailed in Supplementary Tables \ref{tab: roberta base results}-\ref{tab: roberta large results}, indicate that SSB is significantly superior to SSL and (IA)$^3$. Additionally, while training fewer than one-thousandth of the parameters, SSB closely approximates the outcomes of fully fine-tuning.
These findings underscore the effectiveness of the adjustments specified in Eq. (\ref{eq adjust the row}), confirming the potential of subspace scaling.

\subsubsection{Nonlinear Mapping}
Beyond scaling, Mode 2 methods may involve nonlinear transformations, which allows it to capture complex relationships between $\mathbf{W}$ and $\mathbf{W}^{*}$ that are not possible with mere scaling.

\noindent
\textbf{BitFit and its Derivatives:}
BitFit \cite{zaken2021bitfit} is designed to optimize solely the bias terms within a model while keeping all other parameters frozen, achieving performance comparable to full fine-tuning. Extending this concept, S-BitFit \cite{lawton2023neural} integrates Network Architecture Search (NAS) with the BitFit strategy, maintaining the structural integrity of BitFit by imposing constraints on the NAS algorithm to determine whether the gradient of the bias term should be zero.

We consider the scenario of fine-tuning the bias term of a layer. For an input $\mathbf{x}\in\mathbb{R}^{l\times n}$ with frozen weights $\mathbf{W}$ and bias term $\mathbf{b}\in\mathbb{R}^{m}$, the output of the layer is computed as follows:
\begin{equation}
    {\rm output} = \mathbf{xW} + \mathbf{1}_l\mathbf{b}^\mathsf{T},
\end{equation}
where $\mathbf{1}_l\in\mathbb{R}^{l}$ is an all-one vector.
To facilitate the integration of the bias term into the weight matrix, we can augment $\mathbf{W}$ by appending $\mathbf{b}^{\mathsf{T}}$ as an additional row. This alteration leads to the following representation:
\begin{equation}
    {\rm output} = \begin{bmatrix}
        \mathbf{x} & \mathbf{1}
    \end{bmatrix} \begin{bmatrix}
        \mathbf{W}\\
        \mathbf{b}^\mathsf{T}
    \end{bmatrix} = \hat{\mathbf{x}}\hat{\mathbf{W}},
\end{equation}
where $\mathbf{1}\in\mathbb{R}^{l}$
and $\hat{\mathbf{W}}\in\mathbb{R}^{(n+1)\times m}$ is the augmented matrix. Therefore, BitFit fundamentally involves fine-tuning each element of the final row of $\hat{\mathbf{W}}$, corresponding directly to reconstructing the row space of the augmented weight matrix.

\noindent
\textbf{Soft Prompt Derivatives:}
Soft prompt derivatives, such as Prefix-tuning \cite{li2021prefix}, and prompt-tuning \cite{lester2021power}, are prevailing in natural language processing \cite{gao2020making, tan2021msp}. Prefix-tuning introduces trainable continuous tokens, or prefixes, appended to either the input or output of a layer. These prefixes, sourced from a specific parameter matrix, remain trainable while other parameters of the pre-trained model are fixed during training. Conversely, Prompt-tuning simplifies this approach by incorporating soft prompts solely at the input layer. These prompts also originate from an independent parameter matrix and are updated exclusively through gradient descent. Both methods preserve the original model parameters, providing benefits in low-data scenarios and demonstrating potential for generalization across various tasks.

Focusing on the design rather than specific layers to place prefixes, we consider a general case where for an input $\mathbf{x}\in\mathbb{R}^{l\times n}$ and the output $\mathbf{xW}$. $l$ learnable vectors $\mathbf{P}\in\mathbb{R}^{l\times m}$, known as soft prompts, are concatenated in the following formulation:
\begin{equation}
    {\rm concat}(\mathbf{P}, \mathbf{xW}) = \begin{bmatrix}
        \mathbf{P} \\ \mathbf{xW}
    \end{bmatrix}.
    \label{eq soft prompt}
\end{equation}

Similar to the approach used for BitFit, we can augment the weight matrix to restate Eq. (\ref{eq soft prompt}) as 
\begin{equation}
   \begin{bmatrix}
        \mathbf{P} \\ \mathbf{xW}
    \end{bmatrix} =  \begin{bmatrix}
        \mathbf{I} & \mathbf{0}_{l\times n} \\
        \mathbf{0}_{l\times l} & \mathbf{x}
    \end{bmatrix}
    \begin{bmatrix}
        \mathbf{P} \\ \mathbf{W} 
    \end{bmatrix}= \hat{\mathbf{x}}\hat{\mathbf{W}}.
\end{equation}
Here, $\mathbf{I}\in\mathbb{R}^{l\times l}$ is the identity matrix, $\mathbf{0}_{l\times n}\in\mathbb{R}^{l\times n}$ and $\mathbf{0}_{l\times l}\in\mathbb{R}^{l\times }$ are zero matrices, and $\hat{\mathbf{W}}\in\mathbb{R}^{(n+l)\times m}$ is the augmented matrix. 
Thus, soft prompt derivatives essentially involve adjusting the elements of the initial several rows of the augmented weight matrix $\hat{\mathbf{W}}$, thereby reconstructing the original subspace.

\noindent
\textbf{Others:}
There are also methods that adjust the singular vectors by directly modifying elements within the original weight matrix, such as Diff pruning \cite{guo2020parameter}, FishMask \cite{sung2021training}, Fish-Dip \cite{das2023unified}, Xattn Tuning \cite{gheini2021cross}, SPT \cite{he2023sensitivity}, and PaFi \cite{liao2023parameter}, etc.

\section{Subspace Extension}

\begin{figure*}[!ht]
    \centering
    \includegraphics[width=\textwidth]{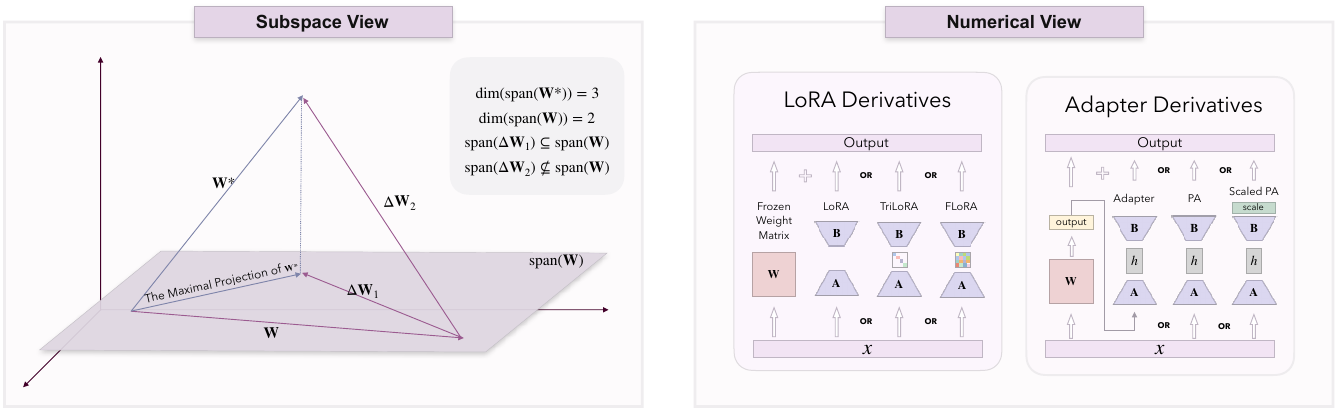}
    \caption{Subspace and Numerical views of extension-based methods. Extension-based methods introduce an additional weight matrix and then try to find the optimal weight projection within the subspace spanned by this additional weight and the original weight. To achieve this, the basis of the subspace constructed by the additional matrix should complement the basis of the original weights as much as possible. The right figure lists some common extension-based methods and their operations on matrices.}
    \label{fig: extension}
\end{figure*}

Extension-based methods aim to approximate the optimal weight matrix $\mathbf{W}^{*}$ by expanding the subspace spanned by the original weight matrix $\mathbf{W} \in \mathbb{R}^{n \times m}$. This is achieved by introducing an additive component $\Delta \mathbf{W} \in \mathbb{R}^{n \times m}$, resulting in an extended subspace that better captures the bases necessary for the target task. It aims to find the closest projection of the optimal weight $\mathbf{W}^{*}$ within this new space (Fig. \ref{fig: extension}).
The transformation function for these methods is defined as:
\begin{equation}
\phi(\mathbf{W}) = g(\mathbf{W}) = \mathbf{W} + s \Delta \mathbf{W},
\end{equation}
where $s \in \mathbb{R}$ is a scaling factor that adjusts the contribution of the additive component. The goal is to find $\Delta \mathbf{W}$ and $s$ such that $\phi(\mathbf{W})$ closely approximates $\mathbf{W}^{}$:
\begin{equation}
\min_{\Delta \mathbf{W}, s} \ \ell\left( \mathbf{W}^{}, \mathbf{W} + s \Delta \mathbf{W} \right),
\end{equation}
where $\ell$ is a loss function, such as the Frobenius norm of the difference.

Assuming $n \leq m$, the column space of $\mathbf{W}$, denoted as $\mathcal{C}(\mathbf{W})$, is a subspace of $\mathbb{R}^n$ with dimension at most $\text{rank}(\mathbf{W}) \leq n$. If $\text{rank}(\mathbf{W}) < n$, then $\mathcal{C}(\mathbf{W})$ does not span $\mathbb{R}^n$. To approximate $\mathbf{W}^{*}$ effectively, it may be necessary to consider bases outside $\mathcal{C}(\mathbf{W})$.
Therefore, the additive component $\Delta \mathbf{W}$ is expected to introduce new bases to the subspace. 
Ideally, the combined column space $\mathcal{C}(\mathbf{W}) + \mathcal{C}(\Delta \mathbf{W})$ should approximate the column space of $\mathbf{W}^{*}$:
\begin{equation}
\mathcal{C}(\mathbf{W}^{*}) \subseteq \mathcal{C}(\mathbf{W}) + \mathcal{C}(\Delta \mathbf{W}).
\end{equation}
Given that we do not have prior knowledge of $\mathcal{C}(\mathbf{W}^{*})$, a conservative approach is to design $\Delta \mathbf{W}$ such that $\mathcal{C}(\mathbf{W}) + \mathcal{C}(\Delta \mathbf{W}) = \mathbb{R}^n$. In the idealist scenario, the column basis vectors of $\Delta\mathbf{W}$ should ideally complement those of  $\mathbf{W}$, implying that the column space of $\mathbf{W}^{*}$ represents the direct sum of these spaces.

Additionally, some studies \cite{hu2021lora, si2024unleashing, zhang2024spectral} suggest that $\mathbf{W}$ is highly correlated to $\mathbf{W}^*$, implying that $\mathbf{W}^{*}$ likely shares a substantial subset of common bases with the subspace of $\mathbf{W}$. Therefore, $\Delta\mathbf{W}$ may only need to account for a small subset of bases absent in $\mathbf{W}$ but present in $\mathbf{W}^{*}$, allowing $\Delta\mathbf{W}$ to be a low-rank matrix \cite{hu2021lora, si2024flora}. Furthermore, empirical research demonstrates that full-parameter fine-tuning of pre-trained models can often be reparameterized into optimizations within a low-dimensional subspace \cite{aghajanyan2020intrinsic, li2018measuring}, indicating that the optimal weights vary within this constrained, low-rank subspace.
This low-rank characteristic of $\Delta\mathbf{W}$ underscores the basis for parameter efficiency in extension-based methods.
Therefore, the optimization problem then becomes:
\begin{equation}
\min_{\Delta \mathbf{W}} \ \| \mathbf{W}^{*} - \mathbf{W} - s \Delta \mathbf{W} \|_F^2 \quad \text{s.t.} \quad \text{rank}(\Delta \mathbf{W}) = r \ll n, m.
\end{equation}

An additional critical aspect is the scaling factor $s$. 
For fixed $\mathbf{W}$ and $\Delta \mathbf{W}$, the optimal $s^{*}$ that minimizes the loss function can be found by solving:
\begin{equation}
s^{*} = \arg\min_{s} \ \| \mathbf{W}^{*} - \mathbf{W} - s \Delta \mathbf{W} \|_F^2.
\end{equation}
This is a univariate quadratic minimization problem, solvable in closed form:
\begin{equation}
s^{} = \frac{\langle \mathbf{W}^{*} - \mathbf{W}, \Delta \mathbf{W} \rangle_F}{\| \Delta \mathbf{W} \|_F^2},
\end{equation}
where $\langle \cdot, \cdot \rangle_F$ denotes the Frobenius inner product. 
However, $s$ is usually set as a hyper-parameter \cite{hu2021lora, zhang2022adaptive, si2024flora}. 
During training, the increment matrix $\Delta \mathbf{W}$ changes continuously, so different values of $s$ can potentially guide $\Delta \mathbf{W}$ along distinct learning trajectories. 
As a result, the choice of $s$ can have a significant—or even critical—impact on model performance, implicitly affecting both the scaling and the direction of updates in $\Delta \mathbf{W}$.

\subsection{LoRA and its Derivatives}\label{sec result lora}
Based on the hypothesis that the update $\Delta \mathbf{W}$ is of low rank, LoRA (Low-Rank Adaptation) introduces $\Delta \mathbf{W}$ as a product of two low-rank matrices:
\begin{equation}
\Delta \mathbf{W} = \mathbf{A} \mathbf{B},
\end{equation}
where $\mathbf{A} \in \mathbb{R}^{n \times r}$ and $\mathbf{B} \in \mathbb{R}^{r \times m}$, with $r \ll \{n, m\}$. The rank constraint ensures parameter efficiency.
Without loss of generality, assuming both $\mathbf{A}$ and $\mathbf{B}$ fully utilize the carrying capacities of their low ranks, i.e., ${\rm rank}(\mathbf{A}) = {\rm rank}(\mathbf{B}) = r$.
The addition term in LoRA aligns with the forms of full rank decomposition \cite{piziak1999full}\footnote{In \cite{peng2024sam}, the authors assert that the formulation of LoRA is based on QR decomposition \cite{francis1961qr,kublanovskaya1962some}, which we believe is incorrect.}. 
Subsequent researches employ different forms of matrix decomposition for $\Delta \mathbf{W}$: 

\begin{itemize}
\item \textbf{TriLoRA} \cite{feng2024trilora} and \textbf{AdaLoRA} \cite{zhang2022adaptive} use a decomposition incorporating a diagonal matrix $\mathbf{D} \in \mathbb{R}^{r \times r}$:
$\Delta \mathbf{W} = \mathbf{A} \mathbf{D} \mathbf{B}$.
\item \textbf{FLoRA} \cite{si2024flora} introduces an arbitrary matrix $\mathbf{M} \in \mathbb{R}^{r \times r}$: $\Delta \mathbf{W} = \mathbf{A} \mathbf{M} \mathbf{B}$.
\end{itemize}

\begin{figure*}[!ht]
    \centering
    \includegraphics[width=\textwidth]{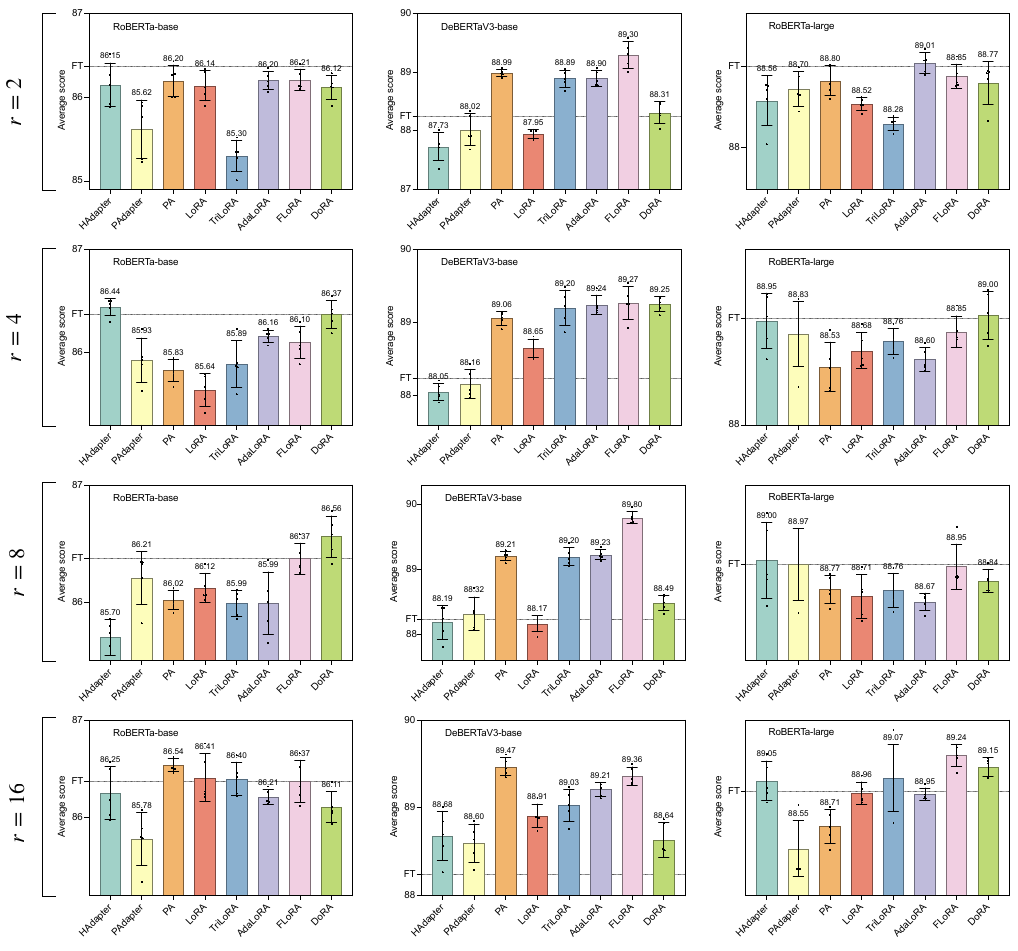}
    \caption{Average score of extension and combination-based methods. Each method is assessed under four different ranks. The horizontal dashed line parallel to the x-axis, labeled FT, represents the performance of fully fine-tuning.  In general the performance of FLoRA is superior to that of AdaLoRA and TriLoRA, followed by LoRA, and the performance of DoRA is superior to that of LoRA.
    The average scores of PEFT methods are evaluated with three large pretrained models, RoBERTa-base \cite{liu2019roberta}, DeBERTaV3-base \cite{he2021debertav3}, and RoBERTa-large \cite{liu2019roberta} on the GLUE benchmark. Each method is assessed under four different ranks 2, 4, 8 and 16. Error bars represent the standard error of the mean across five runs.
    }
    \label{fig: extension result}
\end{figure*}

We first demonstrate that different decomposition forms are capable of being transformed into one another. 
Using the general form of $\Delta\mathbf{W} = \mathbf{AGB}$, we have
\begin{equation}
\begin{aligned}
    & \mathbf{AMB} = \mathbf{{AU}\Sigma {VB}} \ (\mathbf{M} = \mathbf{U\Sigma V}) \\
    = & {\mathbf{A}^{\diamond}\mathbf{D}}\mathbf{B}^{\diamond} \ (\mathbf{A}^{\diamond}=\mathbf{AU}, \mathbf{B}^{\diamond}=\mathbf{VB}) \\ =&  \mathbf{A}^{*}\mathbf{B}^{\diamond} \ \ \ \ (\mathbf{A}^{*}=\mathbf{A}^{\diamond}\mathbf{D}).
\end{aligned}
\label{eq same decomposition formulation}
\end{equation}
Nevertheless, despite their structural similarities, these configurations result in varying levels of effectiveness. 
The effectiveness hierarchy is demonstrated as $\mathcal{P}(\mathbf{AMB}) > \mathcal{P}(\mathbf{ADB}) > \mathcal{P}(\mathbf{AB})$ in general, as evidenced by Fig. \subfig{fig: extension result}{} and studies such as \cite{zhang2022adaptive, hu2021lora, feng2024trilora, si2024flora}. 
Indeed, beyond the methods mentioned, despite the potential for various methods to be theoretically interchangeable in form, there are still performance differences observed between them.
We believe the choice of the $\Delta\mathbf{W}$'s form could influence the feasible set and thus the capacity to approximate $\Delta \mathbf{W}^{*} = \mathbf{W}^{*} - \mathbf{W}$, and thus we formalize this observation with the following proposition:

\begin{proposition}
The expressiveness and optimization landscape of extension-based methods are directly influenced by the decomposition structure of $\Delta \mathbf{W}$, rather than by its mathematical equivalence to other forms.
\label{proposition 1}
\end{proposition}
This proposition is intriguing: If various forms are interchangeable and they all only apply the same low-rank constraint on $\Delta\mathbf{W}$, why do we still see differences in performance? Could a $\Delta\mathbf{W}$ learned by FLoRA for a specific weight not be equivalently learned by other methodologies such as LoRA and TriLoRA? 
This paradox highlights a deeper mystery in extension-based methods — beyond mere theoretical equivalence and the low-rank characteristics traditionally focused on, how exactly does each decomposition form impact the model’s learning dynamics?

To delve into this issue, consider the ideal case where we aim to match $\Delta \mathbf{W}$ to $\Delta \mathbf{W}^{*}$: $\mathbf{A} \mathbf{G} \mathbf{B} = \Delta \mathbf{W}^{*}$.
We utilize SVD on $\Delta\mathbf{W}^{*}$, resulting in $\mathbf{G} = \mathbf{A}^{\dagger} \mathbf{U} \mathbf{\Sigma} \mathbf{V}^T \mathbf{B}^{\dagger}$. 
Suppose $\mathbf{A}^{\dagger} = \mathbf{P}\mathbf{U}^\mathsf{T}$ and $\mathbf{B}^{\dagger} = \mathbf{V}\mathbf{\Sigma}^{\dagger}\mathbf{Q}$,
where $\mathbf{P}\in\mathbb{R}^{r\times n}$ and $\mathbf{Q}\in\mathbb{R}^{n\times r}$ are arbitrary matrices, we can derive $\mathbf{G} = \mathbf{PQ}$.
If $\mathbf{G}$ is an identity matrix, $\mathbf{Q}$ must be the pseudo-inverse of $\mathbf{P}$. 
Building on this setup, if $\mathbf{G}$ is a diagonal matrix, it further specifies that each column of $\mathbf{Q}$, which is the pseudo-inverse of $\mathbf{P}$, may have distinct weights. 
Conversely, when $\mathbf{G}$ is an arbitrary matrix, it does not impose any such constraints.
Therefore, we can arrive at the conclusion: LoRA imposes the most restrictions on the matrix patterns for model learning, followed by AdaLoRA, while FLoRA imposes the least.

At this point, we can explain why different decomposition forms yield varying levels of effectiveness in model training.
Although these methods may have constraints on the matrix patterns, these constraints are not applied during training, which means that LoRA and AdaLoRA are less likely to learn the optimal change compared to FLoRA. 
Further, even if specific matrix patterns are constrained during training, FLoRA’s greater degree of freedom and stronger expressive power in learning make it much easier to achieve better results \cite{hu2021model}. 
This leads us to propose the following:

\begin{proposition}
    Beyond the inherent low-rank constraint of the decomposition itself, different decomposition forms implicitly and subtly impose additional constraints on the matrix patterns. A decomposition that imposes fewer constraints allows for a richer representation of $\Delta \mathbf{W}$, potentially leading to better approximation of $\mathbf{W}^{*}$.
    \label{proposition 2}
\end{proposition}

\begin{figure*}[!ht]
    \centering
    \includegraphics[width=\textwidth]{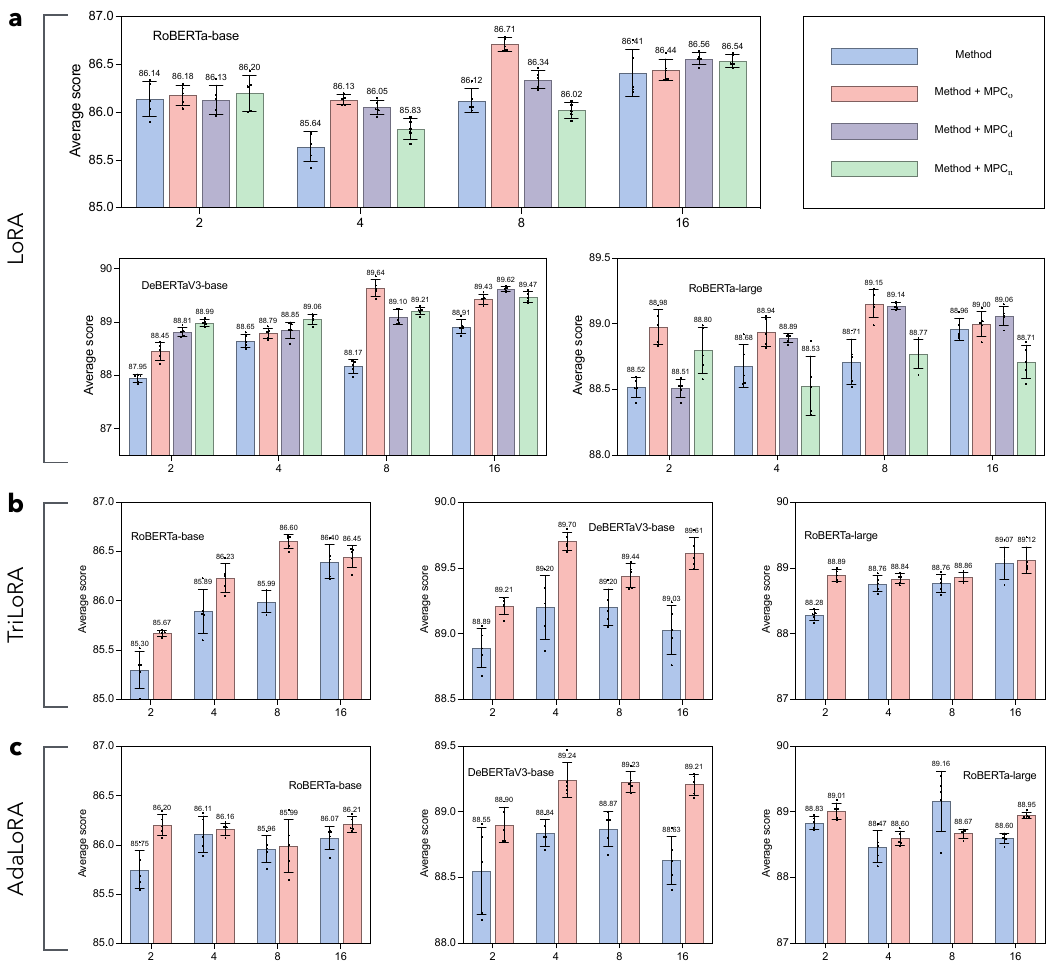}
    \caption{Average score of different methods coupled with MPC framework. \textbf{a}, The performance of LoRA when coupled with MPC$_o$, MPC$_d$, and MPC$_n$. \textbf{b}-\textbf{c}, The performance of TriLoRA and AdaLoRA when coupled with MPC$_o$, respectively. The MPC framework significantly enhances the performance of various PEFT methods, as evaluated with three large pretrained models, RoBERTa-base \cite{liu2019roberta}, DeBERTaV3-base \cite{he2021debertav3}, and RoBERTa-large \cite{liu2019roberta} on the GLUE benchmark. Each method is assessed under four different ranks 2, 4, 8 and 16. Error bars represent the standard error of the mean across five runs.
    }
    \label{fig:MPC}
\end{figure*}

\subsection{Matrix Pattern Constraints (MPC)}
The analysis of different decompositions in the LoRA derivatives provides valuable insights into optimizing model performance through matrix interactions. 
To mitigate the limitations imposed by constrained decompositions, we can introduce regularization terms that encourage desirable properties in $\mathbf{A}$ and $\mathbf{B}$. 
Specifically, we can impose orthogonality or diagonal constraints to enhance the expressiveness while controlling the optimization complexity.

When $\mathbf{G}$ is a diagonal matrix, we simply constrain $\mathbf{A}^{\dagger} = \mathbf{D}_1\mathbf{U}^\mathsf{T}$ and $\mathbf{B}^{\dagger} = \mathbf{V}\mathbf{D}_2$, where $\mathbf{D}_1\in\mathbb{R}^{r\times n}$ and $\mathbf{D}_2\in\mathbb{R}^{m\times r}$ are rectangular diagonal matrices. 
This setup inversely dictates the formulations for $\mathbf{A}$ and $\mathbf{B}$ based on the pseudo-inverses of $\mathbf{D}_1$ and $\mathbf{D}_2$, as $\mathbf{A} = \mathbf{U}\mathbf{D}_1^{\dagger}$ and $\mathbf{B} = \mathbf{D}_2^{\dagger}\mathbf{V}^{\mathsf{T}}$.
This analysis reveals that $\mathbf{A}$ and $\mathbf{B}$ must be structured as linear combinations of the columns of $\mathbf{U}$ and $\mathbf{V}$, respectively. 
Consequently, the matrix relationships are captured as follows:
\begin{equation}
    \begin{aligned}
        \mathbf{A}\mathbf{A}^{\mathsf{T}} = \mathbf{U}\mathbf{D}_1^{\dagger}\mathbf{D}_1^{\dagger\mathsf{T}}\mathbf{U}^{\mathsf{T}}, &  \ \mathbf{A}^{\mathsf{T}}\mathbf{A} = \mathbf{D}_1^{\dagger\mathsf{T}}\mathbf{D}_1^{\dagger},\\
        \mathbf{B}^{\mathsf{T}}\mathbf{B} = \mathbf{V}\mathbf{D}_2^{\dagger\mathsf{T}}\mathbf{D}_2^{\dagger}\mathbf{V}^{\mathsf{T}}, & \ \mathbf{B}\mathbf{B}^{\mathsf{T}} = \mathbf{D}_2^{\dagger}\mathbf{D}_2^{\dagger\mathsf{T}}.
    \end{aligned}
    \label{eq aat bbt}
\end{equation}
By simplifying the constraints on $\mathbf{D}_1 = \mathbf{I}_{r\times n}$ and $\mathbf{D}_2 = \mathbf{I}_{m\times r}$ during training, the following identities hold:
\begin{equation}
\mathbf{D}_1^{\dagger\mathsf{T}}\mathbf{D}_1^{\dagger}=\mathbf{D}_2^{\dagger}\mathbf{D}_2^{\dagger\mathsf{T}} = \mathbf{I}_{r}.
\end{equation}  
In this configuration, $\mathbf{D}_1^{\dagger}\mathbf{D}_1^{\dagger\mathsf{T}}$ and $\mathbf{D}_2^{\dagger\mathsf{T}}\mathbf{D}_2^{\dagger}$ function as block diagonal identity matrices. This arrangement introduces a degree of semi-orthogonality to both $\mathbf{A}$ and $\mathbf{B}$, since
$\mathbf{A}^{\mathsf{T}}\mathbf{A}=\mathbf{B}\mathbf{B}^{\mathsf{T}} =\mathbf{I}_{r}$ and $\mathbf{A}\mathbf{A}^{\mathsf{T}}$ and $\mathbf{B}^\mathsf{T}\mathbf{B}$ are arbitrary matrices. 
Therefore, when $\mathbf{G}$ is a diagonal matrix, we can impose semi-positive definite constraints on the two matrices by introducing regularization term:
\begin{equation}
  \min  \|\mathbf{A}^{\mathsf{T}}\mathbf{A} - \mathbf{I}_r\|_F^2 + \|\mathbf{B}\mathbf{B}^{\mathsf{T}} - \mathbf{I}_r\|_F^2.
  \label{eq adalora constraint orthogonal}
\end{equation}

Extending this further, when $\mathbf{G}$ is an identity matrix, we obtain $\mathbf{I} = \mathbf{A}^{\dagger}\Delta\mathbf{W}^{*}\mathbf{B}^{\dagger}  = \mathbf{D}_1\mathbf{\Sigma}\mathbf{D}_2$.
It becomes evident that $\mathbf{D}_1$ and $\mathbf{D}_2$ are intricately correlated. Therefore, it is feasible to first impose a semi-positive definite constraint to at most one of $\mathbf{A}$ and $\mathbf{B}$, such as $\mathbf{A}$.
For the other matrix $\mathbf{B}$, we can adopt two types of constraints.
One is also constraining $\mathbf{B}$ to exhibit orthogonality, with the regularizer formulated in the Eq. (\ref{eq adalora constraint orthogonal}). 
This constraint assumes that $\mathbf{\Sigma}$ of $\Delta\mathbf{W}^{*}$ is an identity matrix, which makes it a more aggressive approach. 
Alternatively, we can relax the constraint. Eq. (\ref{eq aat bbt}) suggests that $\mathbf{B}^\mathsf{T}\mathbf{B}$ is a diagonal matrix, and hence, we can impose a diagonal constraint on $\mathbf{B}$, with the corresponding regularization term as:
\begin{equation}
     \min \|\mathbf{A}^{\mathsf{T}}\mathbf{A} - \mathbf{I}\|_F^2 + \|\mathbf{B}\mathbf{B}^{\mathsf{T}} - {\rm diag}(\mathbf{B}\mathbf{B}^{\mathsf{T}})\|_F^2.
     \label{eq lora constraint diagonal}
\end{equation}

Furthermore, we can directly break the connection between $\mathbf{D}_1$ and $\mathbf{D}_2$, that is, between $\mathbf{A}$ and $\mathbf{B}$, to alleviate the constraints during the learning process. 
This can be achieved by introducing a non-linear operation between the matrix multiplications of $\mathbf{A}$ and $\mathbf{B}$, effectively breaking the strong link that typically exists between them. 

Overall, we unify the aforementioned three Matrix Pattern Constraints as a novel framework termed as MPC, with MPC$_o$ (Eq. (\ref{eq adalora constraint orthogonal})), MPC$_d$ (Eq. (\ref{eq lora constraint diagonal})) and MPC$_n$ (nonlinear involvement). Without introducing additional parameters, MPC effectively enhance the performance of existing LoRA derivatives, as shown in Figs. \subfig{fig:MPC}{a}-\subfig{fig:MPC}{c} (Supplementary Tables.\ref{tab: roberta base results}-\ref{tab: roberta  large results}). Moreover, MPC can help different methods achieve more stable training. Methods combined with MPC generally exhibit smaller standard deviations compared to those without MPC.

\subsection{Adapter Derivatives}

Adapter derivatives \cite{houlsby2019parameter, pfeiffer2020adapterfusion, he2021towards}, as a pioneering approach in PEFT, integrate small neural modules to adapt models effectively. 
For an input $\mathbf{x}\in\mathbb{R}^{n\times m}$, the adapters are integrated with a residual connection, resulting in the final transformation:
\begin{equation}
    \mathbf{x} \rightarrow \mathbf{x} +  h(\mathbf{xA})\mathbf{B},
    \label{eq adapeter}
\end{equation}
where $h(\cdot)$ is a nonlinear activation function, $\mathbf{A}\in\mathbb{R}^{m\times r}$ and  $\mathbf{B}\in\mathbb{R}^{r\times m}$. This configuration can be further expressed by considering weight matrix as the input $\hat{\mathbf{W}} = \mathbf{x}\in\mathbb{R}^{n\times m}$ and a hypothetical input $\hat{\mathbf{x}} = \mathbf{I}\in\mathbb{R}^{m\times m}$ as 
\begin{equation}
\hat{\mathbf{W}}\hat{\mathbf{x}}\rightarrow \hat{\mathbf{W}}\hat{\mathbf{x}} +h(\hat{\mathbf{W}} \mathbf{A})\mathbf{B}\hat{\mathbf{x}},
    \label{eq adapter trans}
\end{equation}
Therefore, the addition term introduced by the Adapter can be formulated as $\Delta\mathbf{W} = h(\hat{\mathbf{W}}\mathbf{A})\mathbf{B}$, and we can derive
\begin{equation}
    h(\hat{\mathbf{W}}\mathbf{A}) = \Delta\mathbf{W}^{*}\mathbf{B}^{\dagger}.
\end{equation}
Compared with LoRA, where $\mathbf{A} = \Delta\mathbf{W}^{*}\mathbf{B}^{\dagger}$, the Adapter’s inclusion of a nonlinear activation layer further reduces the constraints on learning the relationships between $\mathbf{A}$ and $\mathbf{B}$. According to Proposition \ref{proposition 2}, this reduction in constraints should lead to better model performance. However, the input to the Adapter is fixed as $\mathbf{I}$, which implies that Adapters can only be placed at specific modules within a backbone architecture, potentially limiting their overall effectiveness.

Subsequent developments, such as the MAM Adapter \cite{he2021towards}, which includes the PA or Scaled PA, adopt a parallel architecture similar to LoRA to configure adapter layers. 
Specifically, for an input $\mathbf{x}\in\mathbf{R}^{d\times n}$, we have
\begin{equation}
    \mathbf{x}\mathbf{W} \rightarrow \mathbf{x}\mathbf{W} + h(\mathbf{xA})\mathbf{B},
    \label{eq adapter mpc}
\end{equation}
where $\mathbf{A}\in\mathbb{R}^{n\times r}$ and  $\mathbf{B}\in\mathbb{R}^{r\times m}$. Following Eq. (\ref{eq adapter trans}), we have
\begin{equation}
     \hat{\mathbf{W}}\hat{\mathbf{x}} \rightarrow \hat{\mathbf{W}}\hat{\mathbf{x}}  + h(\hat{\mathbf{W}}\mathbf{A})\mathbf{B},
\end{equation}
resulting in $\Delta\mathbf{W} = h(\hat{\mathbf{W}}\mathbf{A})\mathbf{B}\hat{\mathbf{x}}^{\dagger}$.
This configuration allows Adapters to be applied to any weight matrix within the model. We focus particularly on the design of the PA and its application across various model positions. This nonlinear flexibility typically results in enhanced model performance, as evidenced by our experiments.

\section{Subspace Combination}
Combination-based methods perform both subspace reconstruction and extension simultaneously, blending the principles of both approaches. Moreover, for some methods which can be categorized as both a reconstruction-based and an extension-based method, we also classify them as the combination-based methods. We here analyze several representative combination-based methods as follows.

DoRA \cite{liu2024dora} begins by decomposing the model weights $\mathbf{W}$ into two components: magnitude and direction. The process of adjusting these components is defined as follows:
\begin{equation}
    \phi(\mathbf{W}) = \mathbf{m} \frac{\mathbf{W} + \mathbf{AB}}{\|\mathbf{W} + \mathbf{AB}\|_c},
    \label{eq Dora}
\end{equation}
where $\mathbf{m}\in\mathbb{R}^{1\times m}$ represents the magnitude, and $\|\cdot\|_c$ is the vector-wise norm of a matrix applied across each column. Given that $\mathbf{m}$ is learnable during the training process, this formula can be simplified as:
\begin{equation}
    \phi(\mathbf{W}) = \mathbf{W}\mathbf{D} + \mathbf{ABD},
\end{equation}
where $\mathbf{D}\in\mathbb{R}^{m\times m}$ is a diagonal matrix.
We focus on the extension $\Delta\mathbf{W} = \mathbf{ABD}$ while disregarding the transformation of the column space of $\mathbf{W}$. Following the analysis similar to that for LoRA derivatives, we can derive
\begin{equation}
    \mathbf{I} = \mathbf{A}^{\dagger}\Delta\mathbf{W}^{*}\mathbf{D}^{\dagger}\mathbf{B}^{\dagger} = \mathbf{A}^{\dagger}\mathbf{U}\mathbf{\Sigma}\mathbf{V}^{\mathsf{T}}\mathbf{D}^{\dagger}\mathbf{B}^{\dagger}.
\end{equation}
With the constraints $\mathbf{A}^{\dagger} = \mathbf{D}_1\mathbf{U}^\mathsf{T}$ and $\mathbf{D}^{\dagger}\mathbf{B}^{\dagger} = \mathbf{V}\mathbf{D}_2$, where $\mathbf{D}_1\in\mathbb{R}^{r\times n}$ and $\mathbf{D}_2\in\mathbb{R}^{m\times r}$ are rectangle diagonal, we have 
\begin{equation}
     \mathbf{I} = \mathbf{D}_1\mathbf{\Sigma} \mathbf{D_2}, \ \mathbf{A} = \mathbf{U}\mathbf{D}_1^{\dagger}, \ \mathbf{B}= \mathbf{D}_2^\dagger\mathbf{V}^\mathsf{T}\mathbf{D}^\dagger.
\end{equation}
It is important to note
\begin{equation}
    \mathbf{B}^\mathsf{T}\mathbf{B} = \mathbf{D}^{\dagger\mathsf{T}}\mathbf{V}\mathbf{D}_2^{\dagger\mathsf{T}} \mathbf{D}_2^\dagger\mathbf{V}^\mathsf{T}\mathbf{D}^\dagger, \  \mathbf{B}\mathbf{B}^\mathsf{T} = \mathbf{D}_2^\dagger\mathbf{V}^\mathsf{T}\mathbf{D}^\dagger\mathbf{D}^{\dagger\mathsf{T}}\mathbf{V}\mathbf{D}_2^{\dagger\mathsf{T}}.
\end{equation}
Since ${\rm rank}(\mathbf{D}_2^\dagger) =r$ and the matrix product $\mathbf{D}_2^{\dagger\mathsf{T}} \mathbf{D}_2^\dagger \in \mathbb{R}^{m\times m} $ cannot be an identity matrix, both $\mathbf{B}^\mathsf{T}\mathbf{B}$ and $\mathbf{B}\mathbf{B}^\mathsf{T}$ are arbitrary matrices. 
Therefore, DoRA can at most impose semi-orthogonal constraints on matrix $\mathbf{A}$, while matrix $\mathbf{B}$ remains unconstrained.
Additionally, DoRA reconstructs $\mathbf{W}$ by scaling its column space.
Following an analysis similar to that used for LoRA derivatives, it can be concluded that DoRA may impose a semi-orthogonal constraint on one of the matrices, either $\mathbf{A}$ or $\mathbf{B}$, while leaving the other matrix without any constraints.
Furthermore, DoRA reconstructs $\mathbf{W}$ by scaling its column space. Based on Proposition \ref{proposition 2}, it is concluded that DoRA can lead to superior performance compared to LoRA, as shown in Fig. \ref{fig: extension result} and also noted in \cite{liu2024dora, han2024parameter}.

The Spectral Adapter \cite{zhang2024spectral} is another innovative example within combination-based methods.
Specifically, the Spectral Adapter starts by decomposing the weight matrix $\mathbf{W}$ into its SVD form $\mathbf{W} = \mathbf{U}\mathbf{\Sigma}\mathbf{V}^\mathsf{T}$, and then introduces trainable matrices $\mathbf{A}\in\mathbb{R}^{n\times r}$ and $\mathbf{B}\in\mathbb{R}^{m\times r}$ that are added to the top $r$ columns of the singular vectors $\mathbf{U}$ and $\mathbf{V}$, with the form as 
\begin{equation}
    \phi(\mathbf{W}) = \begin{bmatrix}
        \mathbf{U}_r + \mathbf{A} & \mathbf{U}_{n-r}
    \end{bmatrix} \mathbf{\Sigma}  \begin{bmatrix}
        \mathbf{V}_r + \mathbf{B} & \mathbf{V}_{n-r}
    \end{bmatrix}^{\mathsf{T}},
    \label{eq spectral vector}
\end{equation}
where $\mathbf{U}_r$ and $\mathbf{V}_r$ represent the top $r$ columns of $\mathbf{U}$ and $\mathbf{V}$, and $\mathbf{U}_{n-r}$ and $\mathbf{V}_{n-r}$ account for the remaining columns. The Eq. (\ref{eq spectral vector}) can be rewritten as 
\begin{equation}
    \phi(\mathbf{W}) = \mathbf{W} + \mathbf{A}\mathbf{\Sigma}\mathbf{V}^\mathsf{T} + \mathbf{U}\mathbf{\Sigma}\mathbf{B}^\mathsf{T} + \mathbf{A}\mathbf{\Sigma}\mathbf{B}.
\end{equation}
Therefore, Spectral Adapter can also be viewed as introducing an additional term $\Delta\mathbf{W} = \mathbf{A}\mathbf{\Sigma}\mathbf{V}^\mathsf{T} + \mathbf{U}\mathbf{\Sigma}\mathbf{B}^\mathsf{T} + \mathbf{A}\mathbf{\Sigma}\mathbf{B}$.
This addition is not just a simple reconstruction of the original subspace defined by the singular vectors but also acts as an extension of that subspace.

SVDiff \cite{han2023svdiff} modifies the singular values $\mathbf{\Sigma}$ of the original weight matrix $\mathbf{W}$ by incorporating a diagonal “spectral shift” matrix $\mathbf{D}$. This adjustment is formulated as follows:
\begin{equation}
\phi(\mathbf{W}) = \mathbf{U} h(\mathbf{\Sigma} + \mathbf{D})\mathbf{V}^\mathsf{T},
\label{eq svdiff}
\end{equation}
where $\mathbf{U}$ and $\mathbf{V}$ represent the left and right singular vectors of $\mathbf{W}$, respectively, and $h(\cdot)$ denotes the nonlinear function ReLU. This equation can be expanded to
\begin{equation}
\begin{aligned}
\phi(\mathbf{W}) & = \mathbf{U}\mathbf{\Sigma}\mathbf{V}^\mathsf{T} + \mathbf{U}\ \mathcal{H}_\mathbf{\Sigma}(\mathbf{D})\mathbf{V}^\mathsf{T}\\
& = \mathbf{W} + \Delta\mathbf{W}.
\end{aligned}
\end{equation}
In this context, $\mathcal{H}_\mathbf{A}(\cdot)$ is an element-wise operator where $\mathcal{H}_\mathbf{\Sigma}(\mathbf{D}) = [\max(D{ij}, -\Sigma_{ij})]_{n\times m}$. Consequently, SVDiff not only reconstructs but also extends the subspace. Moreover, we can conclude that the approach, which selectively reconstructs the singular values or vectors by introducing additional trainable components rather than directly altering the singular components, can be categorized as a combination-based method.

\section{Conclusion and Discussion}


The adaptation of pre-trained foundation models for a diverse array of downstream tasks has become a ubiquitous practice in artificial intelligence. Given the extensive range of tasks and the prohibitive costs associated, it is impractical to adjust all parameters comprehensively. In response, the development of parameter-efficient fine-tuning techniques (PEFT) has emerged, facilitating updates to the pre-trained model weights in a manner that is significantly more resource-efficient. 
Although methods of PEFT continue to proliferate, a comprehensive understanding of their underlying mathematical principles and the variance in their performance remains elusive. 
Therefore, in this work, we take the first step by conceptualizing all PEFT methods from a decomposition perspective, unifying them under the subspace tuning methodology. The mathematical foundations underlying each PEFT method are dissected, identifying that each represents a distinct manipulation of the subspace. Inspired by theoretical insights, we propose two novel PEFT methods. Extensive experiments show that by training less than one thousandth of the parameters, can approximate the effects of full fine-tuning. 

Furthermore, we elucidate the reasons behind the performance disparities among different methods. Our analysis yields significant conclusions. The comparative analysis of various PEFT strategies such as LoRA, AdaLoRA, and FLoRA, reveals distinct patterns in their efficacy during training. The more stringent the matrix pattern learning, the more the model performance is constrained. We tested the performance of nearly ten algorithms on three different large pretrained models under four levels of parameter budgets, validating our conclusions with more than 3800 experimental runs. Based on this analysis, we propose a framework that enhances the learning of matrix patterns during model training. The effectiveness of this framework has been confirmed through more than 2000 experimental runs across three methods, four parameter budgets, and three large pretrained models.

The significance of our findings extends beyond the immediate scope of parameter-efficient fine-tuning. The principles underlying PEFT methods can be extrapolated to other domains of artificial intelligence, such as transfer learning \cite{mudrakarta2018k, houlsby2019parameter}, multi-task learning \cite{liu2023pre, mahabadi2021parameter}, and fast training \cite{mahabadi2021parameter,ruckle2020adapterdrop}, and also areas where computational resources are a limiting factor, such as real-time systems \cite{jovanovic2024trends} and embedded devices \cite{feng2023peft}. By analyzing the theoretical aspects of PEFT methods in different scenarios, we can comprehend the underlying logic and, based on these theoretical insights, refine these methods to further enhance their impact across related fields. Additionally, the theoretical underpinnings of subspace tuning present intriguing possibilities for further exploration in this domain as well as others, potentially catalyzing advancements and influencing developments across the broader artificial intelligence landscape.





\bibliographystyle{IEEEtran}
\bibliography{main}

\begin{thebibliography}{10}
\providecommand{\url}[1]{#1}
\csname url@samestyle\endcsname
\providecommand{\newblock}{\relax}
\providecommand{\bibinfo}[2]{#2}
\providecommand{\BIBentrySTDinterwordspacing}{\spaceskip=0pt\relax}
\providecommand{\BIBentryALTinterwordstretchfactor}{4}
\providecommand{\BIBentryALTinterwordspacing}{\spaceskip=\fontdimen2\font plus
\BIBentryALTinterwordstretchfactor\fontdimen3\font minus \fontdimen4\font\relax}
\providecommand{\BIBforeignlanguage}[2]{{%
\expandafter\ifx\csname l@#1\endcsname\relax
\typeout{** WARNING: IEEEtran.bst: No hyphenation pattern has been}%
\typeout{** loaded for the language `#1'. Using the pattern for}%
\typeout{** the default language instead.}%
\else
\language=\csname l@#1\endcsname
\fi
#2}}
\providecommand{\BIBdecl}{\relax}
\BIBdecl

\bibitem{brown2020language}
T.~Brown, B.~Mann, N.~Ryder, M.~Subbiah, J.~D. Kaplan, P.~Dhariwal, A.~Neelakantan, P.~Shyam, G.~Sastry, A.~Askell \emph{et~al.}, ``Language models are few-shot learners,'' \emph{Advances in neural information processing systems}, vol.~33, pp. 1877--1901, 2020.

\bibitem{radford2019language}
A.~Radford, J.~Wu, R.~Child, D.~Luan, D.~Amodei, I.~Sutskever \emph{et~al.}, ``Language models are unsupervised multitask learners,'' \emph{OpenAI blog}, vol.~1, no.~8, p.~9, 2019.

\bibitem{radford2021learning}
A.~Radford, J.~W. Kim, C.~Hallacy, A.~Ramesh, G.~Goh, S.~Agarwal, G.~Sastry, A.~Askell, P.~Mishkin, J.~Clark \emph{et~al.}, ``Learning transferable visual models from natural language supervision,'' in \emph{International conference on machine learning}.\hskip 1em plus 0.5em minus 0.4em\relax PMLR, 2021, pp. 8748--8763.

\bibitem{devlin2018bert}
J.~D. M.-W.~C. Kenton and L.~K. Toutanova, ``Bert: Pre-training of deep bidirectional transformers for language understanding,'' in \emph{Proceedings of naacL-HLT}, vol.~1.\hskip 1em plus 0.5em minus 0.4em\relax Minneapolis, Minnesota, 2019, p.~2.

\bibitem{liu2019roberta}
Y.~Liu, M.~Ott, N.~Goyal, J.~Du, M.~Joshi, D.~Chen, O.~Levy, M.~Lewis, L.~Zettlemoyer, and V.~Stoyanov, ``Roberta: A robustly optimized bert pretraining approach,'' \emph{arXiv preprint arXiv:1907.11692}, 2019.

\bibitem{kirillov2023segment}
A.~Kirillov, E.~Mintun, N.~Ravi, H.~Mao, C.~Rolland, L.~Gustafson, T.~Xiao, S.~Whitehead, A.~C. Berg, W.-Y. Lo \emph{et~al.}, ``Segment anything,'' in \emph{Proceedings of the IEEE/CVF International Conference on Computer Vision}, 2023, pp. 4015--4026.

\bibitem{si2024tendency}
C.~Si, X.~Wang, X.~Yang, and W.~Shen, ``Tendency-driven mutual exclusivity for weakly supervised incremental semantic segmentation,'' in \emph{European Conference on Computer Vision}.\hskip 1em plus 0.5em minus 0.4em\relax Springer, 2025, pp. 37--54.

\bibitem{zhang2023customized}
K.~Zhang and D.~Liu, ``Customized segment anything model for medical image segmentation,'' \emph{arXiv preprint arXiv:2304.13785}, 2023.

\bibitem{zhang2023comprehensive}
C.~Zhang, L.~Liu, Y.~Cui, G.~Huang, W.~Lin, Y.~Yang, and Y.~Hu, ``A comprehensive survey on segment anything model for vision and beyond,'' \emph{arXiv preprint arXiv:2305.08196}, 2023.

\bibitem{achiam2023gpt}
L.~B.~Y. Ai, C.~Ai, and R.~Ai, ``Gpt-4 technical report.''

\bibitem{waisberg2023gpt}
E.~Waisberg, J.~Ong, M.~Masalkhi, S.~A. Kamran, N.~Zaman, P.~Sarker, A.~G. Lee, and A.~Tavakkoli, ``Gpt-4: a new era of artificial intelligence in medicine,'' \emph{Irish Journal of Medical Science (1971-)}, vol. 192, no.~6, pp. 3197--3200, 2023.

\bibitem{mao2023gpteval}
R.~Mao, G.~Chen, X.~Zhang, F.~Guerin, and E.~Cambria, ``Gpteval: A survey on assessments of chatgpt and gpt-4,'' in \emph{Proceedings of the 2024 Joint International Conference on Computational Linguistics, Language Resources and Evaluation (LREC-COLING 2024)}, 2024, pp. 7844--7866.

\bibitem{ma2024segment}
J.~Ma, Y.~He, F.~Li, L.~Han, C.~You, and B.~Wang, ``Segment anything in medical images,'' \emph{Nature Communications}, vol.~15, no.~1, p. 654, 2024.

\bibitem{raffel2020exploring}
C.~Raffel, N.~Shazeer, A.~Roberts, K.~Lee, S.~Narang, M.~Matena, Y.~Zhou, W.~Li, and P.~J. Liu, ``Exploring the limits of transfer learning with a unified text-to-text transformer,'' \emph{Journal of machine learning research}, vol.~21, no. 140, pp. 1--67, 2020.

\bibitem{qiu2020pre}
X.~Qiu, T.~Sun, Y.~Xu, Y.~Shao, N.~Dai, and X.~Huang, ``Pre-trained models for natural language processing: A survey,'' \emph{Science China Technological Sciences}, vol.~63, no.~10, pp. 1872--1897, 2020.

\bibitem{chen2024parameter}
W.~Chen, Z.~Miao, and Q.~Qiu, ``Parameter-efficient tuning of large convolutional models,'' \emph{arXiv preprint arXiv:2403.00269}, 2024.

\bibitem{guo2020parameter}
D.~Guo, A.~Rush, and Y.~Kim, ``Parameter-efficient transfer learning with diff pruning,'' in \emph{Annual Meeting of the Association for Computational Linguistics}, 2021.

\bibitem{he2021towards}
\BIBentryALTinterwordspacing
J.~He, C.~Zhou, X.~Ma, T.~Berg-Kirkpatrick, and G.~Neubig, ``Towards a unified view of parameter-efficient transfer learning,'' in \emph{International Conference on Learning Representations}, 2022. [Online]. Available: \url{https://openreview.net/forum?id=0RDcd5Axok}
\BIBentrySTDinterwordspacing

\bibitem{hu2021lora}
\BIBentryALTinterwordspacing
E.~J. Hu, yelong shen, P.~Wallis, Z.~Allen-Zhu, Y.~Li, S.~Wang, L.~Wang, and W.~Chen, ``Lo{RA}: Low-rank adaptation of large language models,'' in \emph{International Conference on Learning Representations}, 2022. [Online]. Available: \url{https://openreview.net/forum?id=nZeVKeeFYf9}
\BIBentrySTDinterwordspacing

\bibitem{liu2024dora}
S.-y. Liu, C.-Y. Wang, H.~Yin, P.~Molchanov, Y.-C.~F. Wang, K.-T. Cheng, and M.-H. Chen, ``Dora: Weight-decomposed low-rank adaptation,'' in \emph{Forty-first International Conference on Machine Learning}, 2024.

\bibitem{ding2023parameter}
N.~Ding, Y.~Qin, G.~Yang, F.~Wei, Z.~Yang, Y.~Su, S.~Hu, Y.~Chen, C.-M. Chan, W.~Chen \emph{et~al.}, ``Parameter-efficient fine-tuning of large-scale pre-trained language models,'' \emph{Nature Machine Intelligence}, vol.~5, no.~3, pp. 220--235, 2023.

\bibitem{houlsby2019parameter}
N.~Houlsby, A.~Giurgiu, S.~Jastrzebski, B.~Morrone, Q.~De~Laroussilhe, A.~Gesmundo, M.~Attariyan, and S.~Gelly, ``Parameter-efficient transfer learning for nlp,'' in \emph{International conference on machine learning}.\hskip 1em plus 0.5em minus 0.4em\relax PMLR, 2019, pp. 2790--2799.

\bibitem{chen2022adaptformer}
S.~Chen, C.~Ge, Z.~Tong, J.~Wang, Y.~Song, J.~Wang, and P.~Luo, ``Adaptformer: Adapting vision transformers for scalable visual recognition,'' \emph{Advances in Neural Information Processing Systems}, vol.~35, pp. 16\,664--16\,678, 2022.

\bibitem{luo2023towards}
G.~Luo, M.~Huang, Y.~Zhou, X.~Sun, G.~Jiang, Z.~Wang, and R.~Ji, ``Towards efficient visual adaption via structural re-parameterization,'' \emph{arXiv preprint arXiv:2302.08106}, 2023.

\bibitem{mahabadi2021parameter}
R.~K. Mahabadi, S.~Ruder, M.~Dehghani, and J.~Henderson, ``Parameter-efficient multi-task fine-tuning for transformers via shared hypernetworks,'' in \emph{Proceedings of the 59th Annual Meeting of the Association for Computational Linguistics and the 11th International Joint Conference on Natural Language Processing (Volume 1: Long Papers)}, 2021, pp. 565--576.

\bibitem{karimi2021compacter}
R.~Karimi~Mahabadi, J.~Henderson, and S.~Ruder, ``Compacter: Efficient low-rank hypercomplex adapter layers,'' \emph{Advances in Neural Information Processing Systems}, vol.~34, pp. 1022--1035, 2021.

\bibitem{lester2021power}
B.~Lester, R.~Al-Rfou, and N.~Constant, ``The power of scale for parameter-efficient prompt tuning,'' in \emph{Proceedings of the 2021 Conference on Empirical Methods in Natural Language Processing}, 2021, pp. 3045--3059.

\bibitem{razdaibiedina2023residual}
A.~Razdaibiedina, Y.~Mao, M.~Khabsa, M.~Lewis, R.~Hou, J.~Ba, and A.~Almahairi, ``Residual prompt tuning: improving prompt tuning with residual reparameterization,'' in \emph{The 61st Annual Meeting Of The Association For Computational Linguistics}, 2023.

\bibitem{wang2023non}
Y.~Wang, J.~Wu, T.~Dabral, J.~Zhang, G.~Brown, C.-T. Lu, F.~Liu, Y.~Liang, B.~Pang, M.~Bendersky \emph{et~al.}, ``Non-intrusive adaptation: Input-centric parameter-efficient fine-tuning for versatile multimodal modeling,'' \emph{arXiv preprint arXiv:2310.12100}, 2023.

\bibitem{shi2023dept}
\BIBentryALTinterwordspacing
Z.~Shi and A.~Lipani, ``De{PT}: Decomposed prompt tuning for parameter-efficient fine-tuning,'' in \emph{The Twelfth International Conference on Learning Representations}, 2024. [Online]. Available: \url{https://openreview.net/forum?id=KjegfPGRde}
\BIBentrySTDinterwordspacing

\bibitem{fischer2024prompt}
M.~Fischer, A.~Bartler, and B.~Yang, ``Prompt tuning for parameter-efficient medical image segmentation,'' \emph{Medical Image Analysis}, vol.~91, p. 103024, 2024.

\bibitem{hyeon2021fedpara}
\BIBentryALTinterwordspacing
N.~Hyeon-Woo, M.~Ye-Bin, and T.-H. Oh, ``Fedpara: Low-rank hadamard product for communication-efficient federated learning,'' in \emph{International Conference on Learning Representations}, 2022. [Online]. Available: \url{https://openreview.net/forum?id=d71n4ftoCBy}
\BIBentrySTDinterwordspacing

\bibitem{qiu2023controlling}
Z.~Qiu, W.~Liu, H.~Feng, Y.~Xue, Y.~Feng, Z.~Liu, D.~Zhang, A.~Weller, and B.~Sch{\"o}lkopf, ``Controlling text-to-image diffusion by orthogonal finetuning,'' \emph{Advances in Neural Information Processing Systems}, vol.~36, pp. 79\,320--79\,362, 2023.

\bibitem{renduchintala2023tied}
A.~Renduchintala, T.~Konuk, and O.~Kuchaiev, ``Tied-lora: Enhancing parameter efficiency of lora with weight tying,'' in \emph{Proceedings of the 2024 Conference of the North American Chapter of the Association for Computational Linguistics: Human Language Technologies (Volume 1: Long Papers)}, 2024, pp. 8686--8697.

\bibitem{kopiczko2023vera}
\BIBentryALTinterwordspacing
D.~J. Kopiczko, T.~Blankevoort, and Y.~M. Asano, ``Ve{RA}: Vector-based random matrix adaptation,'' in \emph{The Twelfth International Conference on Learning Representations}, 2024. [Online]. Available: \url{https://openreview.net/forum?id=NjNfLdxr3A}
\BIBentrySTDinterwordspacing

\bibitem{yeh2023navigating}
S.-Y. YEH, Y.-G. Hsieh, Z.~Gao, B.~B. Yang, G.~Oh, and Y.~Gong, ``Navigating text-to-image customization: From lycoris fine-tuning to model evaluation,'' in \emph{The Twelfth International Conference on Learning Representations}, 2023.

\bibitem{zhang2022adaptive}
Q.~Zhang, M.~Chen, A.~Bukharin, P.~He, Y.~Cheng, W.~Chen, and T.~Zhao, ``Adaptive budget allocation for parameter-efficient fine-tuning,'' in \emph{The Eleventh International Conference on Learning Representations}, 2022.

\bibitem{si2024flora}
C.~Si, X.~Wang, X.~Yang, Z.~Xu, Q.~Li, J.~Dai, Y.~Qiao, X.~Yang, and W.~Shen, ``Flora: Low-rank core space for n-dimension,'' \emph{arXiv preprint arXiv:2405.14739}, 2024.

\bibitem{si2024unleashing}
C.~Si, Z.~Shi, S.~Zhang, X.~Yang, H.~Pfister, and W.~Shen, ``Unleashing the power of task-specific directions in parameter efficient fine-tuning,'' \emph{arXiv preprint arXiv:2409.01035}, 2024.

\bibitem{zaken2021bitfit}
E.~B. Zaken, Y.~Goldberg, and S.~Ravfogel, ``Bitfit: Simple parameter-efficient fine-tuning for transformer-based masked language-models,'' in \emph{Proceedings of the 60th Annual Meeting of the Association for Computational Linguistics (Volume 2: Short Papers)}, 2022, pp. 1--9.

\bibitem{lawton2023neural}
N.~G. Lawton, A.~Kumar, G.~Thattai, A.~Galstyan, and G.~Ver~Steeg, ``Neural architecture search for parameter-efficient fine-tuning of large pre-trained language models,'' in \emph{The 61st Annual Meeting Of The Association For Computational Linguistics}, 2023.

\bibitem{han2024parameter}
\BIBentryALTinterwordspacing
Z.~Han, C.~Gao, J.~Liu, J.~Zhang, and S.~Q. Zhang, ``Parameter-efficient fine-tuning for large models: A comprehensive survey,'' \emph{Transactions on Machine Learning Research}, 2024. [Online]. Available: \url{https://openreview.net/forum?id=lIsCS8b6zj}
\BIBentrySTDinterwordspacing

\bibitem{fu2023effectiveness}
Z.~Fu, H.~Yang, A.~M.-C. So, W.~Lam, L.~Bing, and N.~Collier, ``On the effectiveness of parameter-efficient fine-tuning,'' in \emph{Proceedings of the AAAI Conference on Artificial Intelligence}, vol.~37, no.~11, 2023, pp. 12\,799--12\,807.

\bibitem{mudrakarta2018k}
P.~K. Mudrakarta, M.~Sandler, A.~Zhmoginov, and A.~Howard, ``K for the price of 1: Parameter-efficient multi-task and transfer learning,'' in \emph{International Conference on Learning Representations}, 2018.

\bibitem{si2024partial}
C.~Si, Z.~Jiang, X.~Wang, Y.~Wang, X.~Yang, and W.~Shen, ``Partial label learning with a partner,'' in \emph{Proceedings of the AAAI Conference on Artificial Intelligence}, vol.~38, no.~13, 2024, pp. 15\,029--15\,037.

\bibitem{jovanovic2024trends}
M.~Jovanovic and P.~Voss, ``Trends and challenges of real-time learning in large language models: A critical review,'' \emph{arXiv preprint arXiv:2404.18311}, 2024.

\bibitem{feng2023peft}
T.~Feng and S.~Narayanan, ``Peft-ser: On the use of parameter efficient transfer learning approaches for speech emotion recognition using pre-trained speech models,'' in \emph{2023 11th International Conference on Affective Computing and Intelligent Interaction (ACII)}.\hskip 1em plus 0.5em minus 0.4em\relax IEEE, 2023, pp. 1--8.

\bibitem{si2023multi}
C.~Si, Y.~Jia, R.~Wang, M.-L. Zhang, Y.~Feng, and Q.~Chongxiao, ``Multi-label classification with high-rank and high-order label correlations,'' \emph{IEEE Transactions on Knowledge and Data Engineering}, 2023.

\bibitem{wang2024loraga}
\BIBentryALTinterwordspacing
S.~Wang, L.~Yu, and J.~Li, ``Lo{RA}-{GA}: Low-rank adaptation with gradient approximation,'' in \emph{The Thirty-eighth Annual Conference on Neural Information Processing Systems}, 2024. [Online]. Available: \url{https://openreview.net/forum?id=VaLAWrLHJv}
\BIBentrySTDinterwordspacing

\bibitem{meng2024pissa}
\BIBentryALTinterwordspacing
F.~Meng, Z.~Wang, and M.~Zhang, ``Pi{SSA}: Principal singular values and singular vectors adaptation of large language models,'' in \emph{The Thirty-eighth Annual Conference on Neural Information Processing Systems}, 2024. [Online]. Available: \url{https://openreview.net/forum?id=6ZBHIEtdP4}
\BIBentrySTDinterwordspacing

\bibitem{wang2024lorapro}
Z.~Wang and J.~Liang, ``Lora-pro: Are low-rank adapters properly optimized?'' \emph{arXiv preprint arXiv:2407.18242}, 2024.

\bibitem{hao2024flora}
Y.~Hao, Y.~Cao, and L.~Mou, ``Flora: Low-rank adapters are secretly gradient compressors,'' in \emph{Forty-first International Conference on Machine Learning}, 2024.

\bibitem{he2021debertav3}
P.~He, J.~Gao, and W.~Chen, ``Debertav3: Improving deberta using electra-style pre-training with gradient-disentangled embedding sharing,'' in \emph{The Eleventh International Conference on Learning Representations}, 2021.

\bibitem{peng2024sam}
Z.~Peng, Z.~Xu, Z.~Zeng, X.~Yang, and W.~Shen, ``Sam-parser: Fine-tuning sam efficiently by parameter space reconstruction,'' in \emph{Proceedings of the AAAI Conference on Artificial Intelligence}, vol.~38, no.~5, 2024, pp. 4515--4523.

\bibitem{liu2022few}
H.~Liu, D.~Tam, M.~Muqeeth, J.~Mohta, T.~Huang, M.~Bansal, and C.~A. Raffel, ``Few-shot parameter-efficient fine-tuning is better and cheaper than in-context learning,'' \emph{Advances in Neural Information Processing Systems}, vol.~35, pp. 1950--1965, 2022.

\bibitem{li2021prefix}
X.~L. Li and P.~Liang, ``Prefix-tuning: Optimizing continuous prompts for generation,'' in \emph{Proceedings of the 59th Annual Meeting of the Association for Computational Linguistics and the 11th International Joint Conference on Natural Language Processing (Volume 1: Long Papers)}, 2021, pp. 4582--4597.

\bibitem{gao2020making}
T.~Gao, A.~Fisch, and D.~Chen, ``Making pre-trained language models better few-shot learners,'' in \emph{Proceedings of the 59th Annual Meeting of the Association for Computational Linguistics and the 11th International Joint Conference on Natural Language Processing (Volume 1: Long Papers)}, 2021, pp. 3816--3830.

\bibitem{tan2021msp}
Z.~Tan, X.~Zhang, S.~Wang, and Y.~Liu, ``Msp: Multi-stage prompting for making pre-trained language models better translators,'' in \emph{Proceedings of the 60th Annual Meeting of the Association for Computational Linguistics (Volume 1: Long Papers)}, 2022, pp. 6131--6142.

\bibitem{sung2021training}
Y.-L. Sung, V.~Nair, and C.~A. Raffel, ``Training neural networks with fixed sparse masks,'' \emph{Advances in Neural Information Processing Systems}, vol.~34, pp. 24\,193--24\,205, 2021.

\bibitem{das2023unified}
S.~S.~S. Das, R.~H. Zhang, P.~Shi, W.~Yin, and R.~Zhang, ``Unified low-resource sequence labeling by sample-aware dynamic sparse finetuning,'' in \emph{2023 Conference on Empirical Methods in Natural Language Processing, EMNLP 2023}.\hskip 1em plus 0.5em minus 0.4em\relax Association for Computational Linguistics (ACL), 2023, pp. 6998--7010.

\bibitem{gheini2021cross}
M.~Gheini, X.~Ren, and J.~May, ``Cross-attention is all you need: Adapting pretrained transformers for machine translation,'' in \emph{Proceedings of the 2021 Conference on Empirical Methods in Natural Language Processing}, 2021, pp. 1754--1765.

\bibitem{he2023sensitivity}
H.~He, J.~Cai, J.~Zhang, D.~Tao, and B.~Zhuang, ``Sensitivity-aware visual parameter-efficient fine-tuning,'' in \emph{Proceedings of the IEEE/CVF International Conference on Computer Vision}, 2023, pp. 11\,825--11\,835.

\bibitem{liao2023parameter}
B.~Liao, Y.~Meng, and C.~Monz, ``Parameter-efficient fine-tuning without introducing new latency,'' in \emph{Proceedings of the 61st Annual Meeting of the Association for Computational Linguistics (Volume 1: Long Papers)}, 2023, pp. 4242--4260.

\bibitem{zhang2024spectral}
\BIBentryALTinterwordspacing
F.~Zhang and M.~Pilanci, ``Spectral adapter: Fine-tuning in spectral space,'' in \emph{The Thirty-eighth Annual Conference on Neural Information Processing Systems}, 2024. [Online]. Available: \url{https://openreview.net/forum?id=UoxuaOGV6B}
\BIBentrySTDinterwordspacing

\bibitem{aghajanyan2020intrinsic}
A.~Aghajanyan, S.~Gupta, and L.~Zettlemoyer, ``Intrinsic dimensionality explains the effectiveness of language model fine-tuning,'' in \emph{Proceedings of the 59th Annual Meeting of the Association for Computational Linguistics and the 11th International Joint Conference on Natural Language Processing (Volume 1: Long Papers)}, 2021, pp. 7319--7328.

\bibitem{li2018measuring}
C.~Li, H.~Farkhoor, R.~Liu, and J.~Yosinski, ``Measuring the intrinsic dimension of objective landscapes,'' in \emph{International Conference on Learning Representations}, 2018.

\bibitem{piziak1999full}
R.~Piziak and P.~L. Odell, ``Full rank factorization of matrices,'' \emph{Mathematics magazine}, vol.~72, no.~3, pp. 193--201, 1999.

\bibitem{francis1961qr}
J.~G. Francis, ``The qr transformation a unitary analogue to the lr transformation—part 1,'' \emph{The Computer Journal}, vol.~4, no.~3, pp. 265--271, 1961.

\bibitem{kublanovskaya1962some}
V.~N. Kublanovskaya, ``On some algorithms for the solution of the complete eigenvalue problem,'' \emph{USSR Computational Mathematics and Mathematical Physics}, vol.~1, no.~3, pp. 637--657, 1962.

\bibitem{feng2024trilora}
C.~Feng, M.~He, Q.~Tian, H.~Yin, X.~Zhao, H.~Tang, and X.~Wei, ``Trilora: Integrating svd for advanced style personalization in text-to-image generation,'' \emph{arXiv preprint arXiv:2405.11236}, 2024.

\bibitem{hu2021model}
X.~Hu, L.~Chu, J.~Pei, W.~Liu, and J.~Bian, ``Model complexity of deep learning: A survey,'' \emph{Knowledge and Information Systems}, vol.~63, pp. 2585--2619, 2021.

\bibitem{pfeiffer2020adapterfusion}
J.~Pfeiffer, A.~Kamath, A.~R{\"u}ckl{\'e}, K.~Cho, and I.~Gurevych, ``Adapterfusion: Non-destructive task composition for transfer learning,'' in \emph{Proceedings of the 16th Conference of the European Chapter of the Association for Computational Linguistics: Main Volume}, 2021, pp. 487--503.

\bibitem{han2023svdiff}
L.~Han, Y.~Li, H.~Zhang, P.~Milanfar, D.~Metaxas, and F.~Yang, ``Svdiff: Compact parameter space for diffusion fine-tuning,'' in \emph{Proceedings of the IEEE/CVF International Conference on Computer Vision}, 2023, pp. 7323--7334.

\bibitem{liu2023pre}
P.~Liu, W.~Yuan, J.~Fu, Z.~Jiang, H.~Hayashi, and G.~Neubig, ``Pre-train, prompt, and predict: A systematic survey of prompting methods in natural language processing,'' \emph{ACM Computing Surveys}, vol.~55, no.~9, pp. 1--35, 2023.

\bibitem{ruckle2020adapterdrop}
A.~R{\"u}ckl{\'e}, G.~Geigle, M.~Glockner, T.~Beck, J.~Pfeiffer, N.~Reimers, and I.~Gurevych, ``Adapterdrop: On the efficiency of adapters in transformers,'' in \emph{Proceedings of the 2021 Conference on Empirical Methods in Natural Language Processing}, 2021, pp. 7930--7946.

\end{thebibliography}

\appendices

\newcounter{rownumc}
\newcommand{\rownumber}{\stepcounter{rownumc}\therownumc}

\begin{table*}[ht]
    \centering
    \setcounter{rownumc}{0} 
    \renewcommand\arraystretch{0.9} 
    \caption {Results with RoBERTa-base \cite{liu2019roberta} fine-tuned on GLUE development set.}
    \resizebox{\textwidth}{!}{
    \begin{tabular}{c l | c| c c c c c c c c |
    c}
    \toprule
         \multirow{2}{*}{\textbf{Row}} & \multirow{2}{*}{\textbf{Method}} &  \multirow{2}{*}{\textperthousand \textbf{Params}} & \textbf{MNLI} & \textbf{SST-2} & \textbf{CoLA} & \textbf{QQP} & \textbf{QNLI} & \textbf{RTE} & \textbf{MRPC} & \textbf{STS-B} & \textbf{All}\\
         & & & Acc & Acc & Mcc & Acc & Acc & Acc & Acc & Corr & Avg. \\ 
         \hline
         
          \rownumber & Fully FT & 1000\textperthousand & 87.62 & 94.84 & 63.58 & 91.87 & 92.80 & 78.80 & 90.20 & 91.23 & 86.37\\ 
         \hline

         \rownumber & SAM-PARSER & 0.44\textperthousand & 54.92 & 85.09 & 39.85 & 75.92 & 70.79 & 59.57 & 74.26 & 60.65 & 65.13 \\ \rowcolor{gray!20}
         
          \rownumber & (IA)$^3$ & 0.44\textperthousand & 84.83 & 94.15 & 60.14 & 87.92 & 90.39 & 76.17 & 87.75 & 90.23 & 83.91 \\

        \rownumber & \textbf{SSL} & 0.22\textperthousand & 83.45 & 93.81 & 56.02 & 87.30 & 89.20 & 74.01 & 86.76 & 89.52 & 82.51 \\  \rowcolor{gray!20}
        
         \rownumber & \textbf{SSB} & 0.66\textperthousand & 85.80 & 94.61 & 60.92 & 88.65 & 91.20 & 78.53 & 87.75 & 90.31 & 84.72 \\
          
         \rownumber & BitFit & 0.82\textperthousand & 85.29 & 94.29 & 59.58 & 88.10 & 90.39 & 78.84 & 88.73 & 90.32 & 84.44 \\ 
        \hline

        \rownumber & HAdapter & 2.50\textperthousand & 87.45 & 94.72 & 63.88 & 90.29 & 92.71 & 80.14 & 89.22 & 90.80 & 86.15 \\  \rowcolor{gray!20}
        
        \rownumber & PAdapter & 2.43\textperthousand & 87.11 & 94.15 & 62.74 & 89.95 & 92.71 & 80.14 & 87.99 & 90.13 & 85.62\\
        
        \rownumber & PA & 2.65\textperthousand & 87.55 & 94.38 & 64.23 & 90.03 & 92.81 & 80.51 & 89.22 & 90.85 & 86.20\\  \rowcolor{gray!20}

       \rownumber & LoRA & 2.65\textperthousand & 87.20 & 94.38 & 65.61 & 89.25 & 92.07 & 81.59 & 87.99 & 91.01 & 86.14 \\ 

         \rownumber & LoRA + MPC$_o$ & 2.65\textperthousand & 86.96 & 94.72 & 64.08 & 89.24 & 92.00 & 81.23 & 90.20 & 91.03 & 86.18 \\  \rowcolor{gray!20}
         
        \rownumber & LoRA + MPC$_d$ & 2.65\textperthousand & 87.03 & 94.50 & 64.24 & 89.19 & 92.13 & 81.51 & 89.46 & 91.00 & 86.13 \\ 

        \rownumber & LoRA + MPC$_n$ & 2.65\textperthousand & 87.55 & 94.38 & 64.23 & 90.03 & 92.81 & 80.51 & 89.22 & 90.85 & 86.20 \\   \rowcolor{gray!20}

         \rownumber & TriLoRA & 2.65\textperthousand & 86.81 & 94.61 & 64.47 & 89.61 & 91.82 & 76.53 & 88.24 & 90.31 & 85.30 \\  
 
         \rownumber & TriLoRA + MPC$_o$ & 2.65\textperthousand & 87.48 & 94.27 & 63.97 & 89.93 & 92.55 & 78.34 & 88.24 & 90.55 & 85.67 \\ \rowcolor{gray!20}

        \rownumber & AdaLoRA - MPC$_o$ & 2.65\textperthousand & 87.51 & 94.15 & 62.35 & 90.14 & 92.97 & 80.51 & 87.75 & 90.62 & 85.75 \\ 
       
         \rownumber & AdaLoRA & 2.65\textperthousand & 87.31 & 94.72 & 64.33 & 89.77 & 92.81 & 81.95 & 88.24 & 90.48 & 86.20 \\   \rowcolor{gray!20}

         \rownumber & FLoRA & 2.65\textperthousand & 87.31 & 94.38 & 64.09 & 89.97 & 92.77 & 82.67 & 87.75 & 90.77 & 86.21\\ \hline
        
         \rownumber & DoRA & 3.32\textperthousand & 86.74 & 94.50 & 66.19 & 90.28 & 91.95 & 79.78 & 88.48 & 91.01 & 86.12  \\
         \midrule

         \rownumber &  HAdapter & 4.87\textperthousand & 87.34 & 94.95 & 63.68 & 90.55 & 92.81 & 81.23 & 89.71 & 91.21 & 86.44 \\ \rowcolor{gray!20}
        
         \rownumber & PAdapter & 4.80\textperthousand & 87.22 & 94.84 & 64.91 & 89.95 & 92.38 & 79.06 & 88.75 & 90.31 & 85.93 \\ 

         \rownumber &  PA & 5.31\textperthousand & 87.32 & 94.27 & 62.43 & 90.26 & 92.68 & 79.42 & 89.22 & 91.07 & 85.83\\  \rowcolor{gray!20}

         \rownumber & LoRA & 5.31\textperthousand & 87.13 & 94.38 & 62.52 & 89.40 & 92.20 & 79.42 & 88.97 & 91.11 & 85.64 \\ 

         \rownumber & LoRA + MPC$_o$ & 5.31\textperthousand & 87.02 & 94.38 & 64.58 & 89.47 & 92.33 & 81.23 & 88.97 & 91.02 & 86.13\\   \rowcolor{gray!20}

         \rownumber & LoRA + MPC$_d$ & 5.31\textperthousand & 87.15 & 94.27 & 63.68 & 89.44 & 92.44 & 81.59 & 88.73 & 91.08 & 86.05 \\
        
        \rownumber &  LoRA + MPC$_n$ & 5.31\textperthousand & 87.32 & 94.27 & 62.43 & 90.26 & 92.68 & 79.42 & 89.22 & 91.07 & 85.83\\  \rowcolor{gray!20}
        
        \rownumber & TriLoRA & 5.31\textperthousand & 87.45 & 93.81 & 63.85 & 90.30 & 92.42 & 80.14 & 88.97 & 90.17 & 85.89 \\  

        \rownumber & TriLoRA + MPC$_o$ & 5.31\textperthousand & 87.90 & 94.84 & 62.20 & 90.24 & 92.60 & 83.03 & 88.73 & 90.26 & 86.23 \\ \rowcolor{gray!20}

        \rownumber & AdaLoRA - MPC$_o$ & 5.31\textperthousand & 87.39 & 94.50 & 61.84 & 90.12 & 92.64 & 82.31 & 89.71 & 90.34 & 86.11 \\ 

        \rownumber & AdaLoRA & 5.31\textperthousand & 87.43 & 94.50 & 61.74 & 89.76 & 92.90 & 82.67 & 89.71 & 90.55 & 86.16 \\ \rowcolor{gray!20}

        \rownumber & FLoRA & 5.31\textperthousand & 87.59 & 93.92 & 62.13 & 90.28 & 92.73 & 82.67 & 88.73 & 90.78 & 86.10\\ \hline

        \rownumber & DoRA & 5.97\textperthousand & 87.06 & 94.38 & 66.19 & 90.67 & 92.15 & 80.87 & 88.73 & 90.94 & 86.37  \\

        \midrule

         \rownumber & HAdapter & 9.59\textperthousand & 87.74 & 94.04 & 61.53 & 89.27 & 92.57 & 80.14 & 89.46 & 90.84 & 85.70\\  \rowcolor{gray!20}

         \rownumber & PAdapter & 9.52\textperthousand & 86.90 & 94.50 & 66.52 & 90.05 & 92.46 & 80.87 & 88.24 & 90.13 & 86.21 \\

         \rownumber & PA & 10.62\textperthousand & 87.50 & 94.61 & 62.89 & 89.51 & 92.73 & 80.87 & 89.46 & 90.59 & 86.02 \\  \rowcolor{gray!20}

        \rownumber & LoRA & 10.62\textperthousand & 87.38 & 94.84 & 64.58 & 89.39 & 92.47 & 80.14 & 89.21 & 90.96 & 86.12 \\
        
         \rownumber & LoRA + MPC$_o$ & 10.62\textperthousand & 87.50 & 94.61 & 65.38 & 89.55 & 93.19 & 83.03 & 89.22 & 91.21 & 86.71 \\ \rowcolor{gray!20}
         
         \rownumber & LoRA + MPC$_d$ & 10.62\textperthousand & 87.31 & 94.84 & 64.43 & 89.41 & 92.57 & 82.31 & 88.73 & 91.09 & 86.34 \\
        
          \rownumber & LoRA + MPC$_n$ & 10.62\textperthousand & 87.50 & 94.61 & 62.89 & 89.51 & 92.73 & 80.87 & 89.46 & 90.59 & 86.02 \\ \rowcolor{gray!20}

        \rownumber & TriLoRA & 10.62\textperthousand & 87.44 & 94.84 & 62.91 & 90.70 & 92.82 & 80.14 & 88.73 & 90.37 & 85.99 \\ 

        \rownumber & TriLoRA + MPC$_o$ & 10.62\textperthousand & 87.97 & 94.72 & 65.06 & 90.38 & 93.01 & 81.95 & 89.22 & 90.51 & 86.60 \\ \rowcolor{gray!20}

        \rownumber & AdaLoRA - MPC$_o$ & 10.62\textperthousand & 87.31 & 94.61 & 62.45 & 89.41 & 92.88 & 81.74 & 88.73 & 90.51 & 85.96 \\ 
        
        \rownumber & AdaLoRA & 10.62\textperthousand & 87.13 & 94.61 & 62.61 & 89.49 & 92.62 & 81.95 & 88.97 & 90.53 & 85.99 \\ \rowcolor{gray!20}

        \rownumber & FLoRA & 10.64\textperthousand & 87.43 & 94.27 & 63.31 & 90.38 & 92.75 & 81.59 & 90.44 & 90.82 & 86.37 \\ \hline        

         \rownumber & DoRA & 11.28\textperthousand & 87.14 & 94.50 & 66.06 & 91.02 & 92.13 & 81.95 & 88.73 & 90.96 & 86.56 \\

        \midrule

        \rownumber &  HAdapter & 19.03\textperthousand & 87.31 & 94.72 & 63.91 & 90.80 & 92.49 & 79.78 & 89.95 & 91.00 & 86.25 \\  \rowcolor{gray!20}
        
         \rownumber & PAdapter & 18.96\textperthousand & 86.64 & 94.84 & 64.55 & 90.22 & 92.73 & 76.17 & 90.44 & 90.66 & 85.78 \\  

         \rownumber & PA & 21.24\textperthousand & 87.55 & 94.50 & 65.79 & 90.93 & 92.68 & 80.87 & 89.22 & 90.74 & 86.54 \\  \rowcolor{gray!20}

         \rownumber & LoRA & 21.24\textperthousand & 87.49 & 94.27 & 64.70 & 89.39 & 92.33 & 81.59 & 90.44 & 91.03 & 86.41\\ 

         \rownumber & LoRA + MPC$_o$ & 21.24\textperthousand & 87.44 & 94.95 & 62.62 & 89.48 & 93.15 & 82.67 & 90.20 & 90.99 & 86.44  \\   \rowcolor{gray!20}

         \rownumber & LoRA + MPC$_d$ & 21.24\textperthousand & 87.34 & 94.50 & 65.33 & 89.26 & 92.82 & 83.03 & 88.97 & 91.22 & 86.56 \\ 

          \rownumber & LoRA + MPC$_n$ & 21.24\textperthousand & 87.55 & 94.50 & 65.79 & 90.93 & 92.68 & 80.87 & 89.22 & 90.74 & 86.54 \\ \rowcolor{gray!20}

        \rownumber & TriLoRA & 21.24\textperthousand & 87.54 & 94.50 & 64.72 & 90.98 & 92.93 & 80.87 & 88.73 & 90.93 & 86.40 \\ 

        \rownumber & TriLoRA + MPC$_o$ & 21.24\textperthousand & 88.21 & 94.61 & 62.89 & 90.53 & 93.01 & 82.67 & 88.97 & 90.67 & 86.45 \\ \rowcolor{gray!20}

        \rownumber & AdaLoRA - MPC$_o$ & 21.24\textperthousand & 87.32 & 94.61 & 62.41 & 90.19 & 92.90 & 81.59 & 88.97 & 90.57 & 86.07 \\ 
        
        \rownumber & AdaLoRA & 21.24\textperthousand & 87.33 & 94.50 & 63.91 & 89.30 & 92.62 & 82.31 & 88.97 & 90.75 & 86.21 \\  \rowcolor{gray!20}

        \rownumber & FLoRA & 21.38\textperthousand & 87.64 & 94.95 & 65.17 & 90.90 & 92.53 & 79.78 & 89.71 & 90.75 & 86.37 \\ \hline

        \rownumber & DoRA & 21.90\textperthousand & 86.98 & 94.61 & 65.30 & 90.92 & 92.90 & 78.70 & 88.48 & 91.00 & 86.11 \\
         
        \bottomrule
        
    \end{tabular}
    }
    \label{tab: roberta base results}
\end{table*}

\newpage

\begin{table*}[ht]
    \centering
    \setcounter{rownumc}{0} 
    \renewcommand\arraystretch{0.9} 
    \caption {Results with DeBERTaV3-base \cite{he2021debertav3} fine-tuned on GLUE development set.}
    \resizebox{\textwidth}{!}{
    \begin{tabular}{c l | c| c c c c c c c c |
    c}
    \toprule
         \multirow{2}{*}{\textbf{Row}} & \multirow{2}{*}{\textbf{Method}} &  \multirow{2}{*}{\textperthousand \textbf{Params}} & \textbf{MNLI} & \textbf{SST-2} & \textbf{CoLA} & \textbf{QQP} & \textbf{QNLI} & \textbf{RTE} & \textbf{MRPC} & \textbf{STS-B} & \textbf{All}\\
         & & & Acc & Acc & Mcc & Acc & Acc & Acc & Acc & Corr & Avg. \\ 
         \hline
         
          \rownumber & Fully FT & 1000\textperthousand & 89.90 & 95.63 & 69.19 & 92.40 & 94.03 & 83.75 & 89.46 & 91.60 & 88.24\\ 
         \hline

         \rownumber & SAM-PARSER & 0.30\textperthousand & 68.32 & 84.28 & 55.21 & 81.91 & 75.86 & 66.06 & 76.47 & 82.82 & 73.87\\ \rowcolor{gray!20}

         \rownumber & (IA)$^3$ &  0.30\textperthousand & 89.44 & 95.52 & 67.01 & 89.01 & 91.80 & 79.42 & 88.23 & 90.79 & 86.40 \\  
         
          \rownumber & \textbf{SSL} & 0.15\textperthousand & 88.35 & 95.07 & 66.64 & 88.19 & 90.10 & 82.31 & 88.68 & 90.13 & 86.18\\\rowcolor{gray!20}
        
         \rownumber & \textbf{SSB} & 0.45\textperthousand & 89.86 & 95.53 & 67.82 & 89.87 & 93.41 & 83.75 & 88.72 & 90.94 & 87.49\\
          
         \rownumber & BitFit & 0.54\textperthousand & 89.37 & 94.84 & 66.96 & 88.41 & 92.24 & 78.80 & 87.75 & 91.35 & 86.21\\ 
         
         
        \hline
         
        \rownumber & HAdapter & 1.68\textperthousand & 90.10 & 95.41 & 67.65 & 91.19 & 93.52 & 83.39 & 89.25 & 91.31 & 87.73\\\rowcolor{gray!20}
        
        \rownumber & PAdapter & 1.63\textperthousand & 89.89 & 94.72 & 69.06 & 91.05 & 93.87 & 84.48 & 89.71 & 91.38 & 88.02\\
        
        \rownumber & PA & 1.79\textperthousand & 90.24 & 95.64 & 71.24 & 91.40 & 94.25 & 87.36 & 90.20 & 91.57 & 88.99 \\ \rowcolor{gray!20}

        \rownumber & LoRA & 1.79\textperthousand & 90.03 & 93.92 & 69.15 & 90.61 & 93.37 & 85.56 & 90.19 & 90.75 & 87.95 \\  

         \rownumber & LoRA + MPC$_o$ & 1.79\textperthousand & 89.99 & 93.61 & 69.93 & 91.25 & 94.98 & 86.64 & 89.70 & 91.53 & 88.45 \\ \rowcolor{gray!20}
         
        \rownumber & LoRA + MPC$_d$ & 1.79\textperthousand & 90.06 & 95.30 & 69.91 & 91.29 & 94.34 & 88.08 & 89.95 & 91.53 & 88.81 \\ 

         \rownumber & LoRA + MPC$_n$ & 1.79\textperthousand & 90.24 & 95.64 & 71.24 & 91.40 & 94.25 & 87.36 & 90.20 & 91.57 & 88.99 \\ \rowcolor{gray!20}

         \rownumber & TriLoRA & 1.79\textperthousand & 90.32 & 95.87 & 69.61 & 91.38 & 94.25 & 87.00 & 91.17 & 91.48 & 88.89 \\ 

         \rownumber & TriLoRA + MPC$_o$ & 1.79\textperthousand & 90.39 & 95.87 & 70.00 & 91.38 & 94.25 & 88.09 & 92.16 & 91.53 & 89.21  \\ \rowcolor{gray!20}

        \rownumber & AdaLoRA - MPC$_o$ & 1.79\textperthousand & 90.34 & 95.10 & 69.51 & 90.98 & 94.10 & 87.01 & 89.95 & 91.40 & 88.55  \\ 
       
         \rownumber & AdaLoRA & 1.79\textperthousand & 90.40 & 95.80 & 69.98 & 91.43 & 94.23 & 87.36 & 90.43 & 91.63 & 88.90 \\   \rowcolor{gray!20}

         \rownumber & FLoRA & 1.79\textperthousand & 90.60 & 96.00 & 70.20 & 91.40 & 94.46 & 88.81 & 90.93 & 91.96 & 89.30\\ \hline
        
         \rownumber & DoRA & 2.23\textperthousand & 90.21 & 94.38 & 69.33 & 90.84 & 93.26 & 86.94 & 90.19 & 91.34 & 88.31\\
         \midrule

         \rownumber &  HAdapter & 3.32\textperthousand & 90.12 & 95.30 & 67.87 & 91.30 & 93.76 & 85.56 & 89.22 & 91.30 & 88.05\\\rowcolor{gray!20}
        
         \rownumber & PAdapter & 3.26\textperthousand & 90.15 & 95.53 & 69.48 & 91.27 & 93.98 & 84.12 & 89.22 & 91.52 & 88.16 \\ 

         \rownumber &  PA & 3.61\textperthousand & 90.23 & 96.11 & 69.78 & 91.77 & 94.27 & 88.08 & 90.20 & 92.03 & 89.06 \\\rowcolor{gray!20}

         \rownumber & LoRA & 3.61\textperthousand & 89.90 & 94.15 & 68.87 & 91.16 & 93.59 & 89.89 & 90.19 & 91.46 & 88.65 \\ 

         \rownumber & LoRA + MPC$_o$& 3.61\textperthousand & 89.92 & 95.06 & 69.55 & 92.10 & 93.66 & 87.72 & 90.93 & 91.34 & 88.79 \\ \rowcolor{gray!20}

         \rownumber & LoRA + MPC$_d$ & 3.61\textperthousand &  89.93 & 95.18 & 69.82 & 92.13 & 93.92 & 87.73 & 90.44 & 91.65 & 88.85 \\
        
         \rownumber &  LoRA + MPC$_n$ & 3.61\textperthousand & 90.23 & 96.11 & 69.78 & 91.77 & 94.27 & 88.08 & 90.20 & 92.03 & 89.06 \\ \rowcolor{gray!20}
         
        \rownumber & TriLoRA & 3.61\textperthousand & 90.37 & 96.10 & 69.94 & 91.88 & 94.51 & 88.45 & 90.93 & 91.42 & 89.20\\ 

        \rownumber & TriLoRA + MPC$_o$& 3.61\textperthousand & 90.55 & 96.10 & 72.37 & 91.93 & 94.55 & 88.45 & 92.16 & 91.52 & 89.70 \\ \rowcolor{gray!20}

        \rownumber & AdaLoRA - MPC$_o$& 3.61\textperthousand & 90.40 & 95.76 & 69.86 & 91.40 & 94.53 & 87.00 & 90.20 & 91.58 & 88.84 \\ 

        \rownumber & AdaLoRA & 3.61\textperthousand & 90.27 & 95.76 & 69.58 & 91.47 & 94.65 & 89.53 & 91.42 & 91.23 & 89.24 \\  \rowcolor{gray!20}

        \rownumber & FLoRA & 3.61\textperthousand & 90.45 & 95.76 & 70.72 & 91.44 & 94.20 & 89.27 & 90.93 & 91.35 & 89.27  \\ \hline

        \rownumber & DoRA & 4.06\textperthousand & 90.12 & 96.22 & 68.24 & 91.55 & 94.65 & 89.53 & 92.40 & 91.29 & 89.25 \\

        \midrule

         \rownumber & HAdapter & 6.63\textperthousand &  90.13 & 95.53 & 68.64 & 91.27 & 94.11 & 84.48 & 89.95 & 91.48 & 88.19 \\\rowcolor{gray!20}

         \rownumber & PAdapter & 6.41\textperthousand & 90.33 & 95.61 & 68.77 & 91.40 & 94.29 & 85.20 & 89.46 & 91.54 & 88.32\\
         
        \rownumber & PA & 7.23\textperthousand & 90.18 & 95.99 & 70.69 & 91.80 & 94.34 & 88.81 & 90.20 & 91.68 & 89.21 \\ \rowcolor{gray!20}

        \rownumber & LoRA & 7.23\textperthousand & 89.80 & 93.69 & 69.30 & 91.78 & 92.97 & 85.70 & 90.68 & 91.62 & 88.17 \\

         \rownumber & LoRA + MPC$_o$& 7.23\textperthousand & 90.60 & 95.64 & 72.19 & 92.29 & 93.75 & 89.53 & 91.17 & 91.91 & 89.64 \\ \rowcolor{gray!20} 

         \rownumber & LoRA + MPC$_d$ & 7.23\textperthousand & 90.20 & 95.53 & 70.06 & 92.23 & 93.78 & 88.81 & 90.20 & 91.99 & 89.10 \\ 
        
        \rownumber & LoRA + MPC$_n$ & 7.23\textperthousand & 90.18 & 95.99 & 70.69 & 91.80 & 94.34 & 88.81 & 90.20 & 91.68 & 89.21 \\  \rowcolor{gray!20} 
        
        \rownumber & TriLoRA & 7.23\textperthousand & 90.28 & 95.99 & 68.28 & 91.97 & 94.47 & 89.53 & 91.67 & 91.43 & 89.20\\ 

        \rownumber & TriLoRA + MPC$_o$& 7.23\textperthousand & 90.60 & 96.10 & 70.78 & 91.96 & 94.51 & 89.17 & 90.93 & 91.49 & 89.44 \\  \rowcolor{gray!20} 

        \rownumber & AdaLoRA - MPC$_o$& 7.23\textperthousand & 90.38 & 95.99 & 69.12 & 91.36 & 94.23 & 87.72 & 90.93 &  91.20 & 88.87 \\ 

        \rownumber & AdaLoRA & 7.23\textperthousand & 90.38 & 95.87 & 71.45 & 91.19 & 94.36 & 88.09 & 90.69 & 91.84 & 89.23 \\  \rowcolor{gray!20} 
        
        \rownumber & FLoRA & 7.24\textperthousand & 90.82 & 96.21 & 72.05 & 91.94 & 94.60 & 89.53 & 91.18 & 92.04 & 89.80 \\ \hline        

         \rownumber & DoRA & 7.66\textperthousand & 89.67 & 94.61 & 69.08 & 91.80 & 93.23 & 87.33 & 90.68 & 91.73 & 88.49 \\

        \midrule

        \rownumber &  HAdapter & 12.93\textperthousand & 89.81 & 95.41 & 69.96 & 92.25 & 94.09 & 86.28 & 89.70 & 91.93 & 88.68\\ \rowcolor{gray!20}
        
         \rownumber & PAdapter & 12.88\textperthousand & 89.82 & 94.84 & 70.69 & 92.31 & 94.09 & 86.28 & 89.21 & 91.54 & 88.60\\ 

         \rownumber & PA & 14.43\textperthousand & 90.35 & 95.64 & 71.45 & 92.30 & 94.38 & 88.81 & 90.93 & 91.89 & 89.47\\ \rowcolor{gray!20}

          \rownumber & LoRA & 14.43\textperthousand & 89.49 & 93.78 & 71.61 & 92.33 & 93.92 & 87.36 & 91.17 & 91.63 & 88.91 \\ 

         \rownumber & LoRA + MPC$_o$& 14.43\textperthousand & 90.32 & 95.98 & 72.26 & 92.30 & 93.73 & 88.08 & 91.17 & 91.61 & 89.43 \\  \rowcolor{gray!20}

         \rownumber & LoRA + MPC$_d$ & 14.43\textperthousand & 90.33 & 95.64 & 72.99 & 92.51 & 94.16 & 89.53 & 90.20 & 91.61 & 89.62 \\ 
            
        \rownumber & LoRA + MPC$_n$ & 14.43\textperthousand & 90.35 & 95.64 & 71.45 & 92.30 & 94.38 & 88.81 & 90.93 & 91.89 & 89.47 \\ \rowcolor{gray!20}
        
        \rownumber & TriLoRA & 14.43\textperthousand & 90.20 & 95.87 & 68.64 & 92.25  & 94.36 & 88.81 & 90.69 & 91.39 & 89.03  \\

        \rownumber & TriLoRA + MPC$_o$& 14.43\textperthousand & 90.59 & 96.22 & 72.71 & 92.15 & 94.62 & 88.08 & 90.93 & 91.56 & 89.61 \\ \rowcolor{gray!20}

        \rownumber & AdaLoRA - MPC$_o$& 14.43\textperthousand & 90.23 & 96.10 & 68.49 & 90.72 & 94.31 & 87.72 & 90.20 & 91.30 & 88.63 \\ 

        \rownumber & AdaLoRA & 14.43\textperthousand & 90.21 & 95.87 & 72.83 & 90.87 & 94.47 & 87.72 & 90.44 & 91.23 & 89.21 \\  \rowcolor{gray!20}

        \rownumber & FLoRA & 14.52\textperthousand & 90.09 & 95.87 & 70.71 & 91.67 & 94.47 & 89.53 & 90.93 & 91.58 & 89.36  \\ \hline

        \rownumber & DoRA & 14.87\textperthousand & 90.17 & 93.92 & 69.92 & 92.47 & 93.21 & 87.00 & 90.68 & 91.73 & 88.64\\
         
        \bottomrule
        
    \end{tabular}
    }
    \label{tab: deberta results}
\end{table*}

\newpage

\begin{table*}[ht]
    \centering
    \setcounter{rownumc}{0} 
    \renewcommand\arraystretch{0.9} 
    \caption {Results with RoBERTa-large \cite{liu2019roberta} fine-tuned on GLUE development set.}
    \resizebox{\textwidth}{!}{
    \begin{tabular}{c l | c| c c c c c c c c |
    c}
    \toprule
         \multirow{2}{*}{\textbf{Row}} & \multirow{2}{*}{\textbf{Method}} &  \multirow{2}{*}{\textperthousand \textbf{Params}} & \textbf{MNLI} & \textbf{SST-2} & \textbf{CoLA} & \textbf{QQP} & \textbf{QNLI} & \textbf{RTE} & \textbf{MRPC} & \textbf{STS-B} & \textbf{All}\\
         & & & Acc & Acc & Mcc & Acc & Acc & Acc & Acc & Corr & Avg. \\ 
         \hline
         
          \rownumber & Fully FT & 1000\textperthousand & 90.29 & 96.41 & 68.02 & 92.24 & 94.74 & 86.66 & 90.92 & 92.44 & 88.97\\ 
         \hline

         \rownumber & SAM-PARSER & 0.42\textperthousand & 52.43 & 87.50 & 42.24 & 76.28 & 66.47 & 60.29 & 75.25 & 54.27 & 64.34 \\ \rowcolor{gray!20}
         
          \rownumber & (IA)$^3$ & 0.42\textperthousand & 90.11 & 95.99 & 67.01 & 89.55 & 93.28 & 89.17 & 88.97 & 88.60 & 87.84 \\

        \rownumber & \textbf{SSL} & 0.21\textperthousand & 89.55 & 95.87 & 66.92 & 89.07 & 92.24 & 86.64 & 87.50 & 86.19 & 86.75 \\  \rowcolor{gray!20}
        
         \rownumber & \textbf{SSB} & 0.62\textperthousand & 90.38 & 95.64 & 67.26 & 90.19 & 93.78 & 88.45 & 89.46 & 89.82 & 88.12 \\
          
         \rownumber & BitFit & 0.76\textperthousand & 90.15 & 96.22 & 68.53 & 89.50 & 94.12 & 84.15 & 89.95 & 91.68 & 88.04\\ 
        \hline

        \rownumber & HAdapter & 2.35\textperthousand & 90.66 & 96.22 & 67.25 & 90.82 & 94.78 & 86.28 & 90.20 & 92.30 & 88.56 \\ \rowcolor{gray!20}
        
        \rownumber & PAdapter & 2.29\textperthousand & 90.39 & 96.22 & 67.25 & 90.47 & 94.47 & 88.44 & 89.95 & 92.38 & 88.70\\
        
        \rownumber & PA & 2.49\textperthousand & 90.79 & 96.33 & 66.91 & 90.91 & 94.80 & 87.73 & 90.44 & 92.46 & 88.80 \\  \rowcolor{gray!20}

        \rownumber & LoRA & 2.49\textperthousand & 90.41 & 95.99 & 67.83 & 90.75 & 94.05 & 87.72 & 89.46 & 91.92 & 88.52 \\  

         \rownumber & LoRA + MPC$_o$& 2.49\textperthousand & 90.73 & 96.33 & 68.10 & 91.01 & 94.55 & 88.80 & 89.95 & 92.36 & 88.98 \\  \rowcolor{gray!20}
         
        \rownumber & LoRA + MPC$_d$ & 2.49\textperthousand & 90.50 & 96.33 & 65.29 & 91.15 & 94.60 & 86.64 & 91.18 & 92.38 & 88.51 \\ 

        \rownumber & LoRA + MPC$_n$ & 2.49\textperthousand & 90.79 & 96.33 & 66.91 & 90.91 & 94.80 & 87.73 & 90.44 & 92.46 & 88.80 \\  \rowcolor{gray!20}

         \rownumber & TriLoRA & 2.49\textperthousand & 90.12 & 95.87 & 67.69 & 90.58 & 94.69 & 85.20 & 90.69 & 91.43 & 88.28\\ 

         \rownumber & TriLoRA + MPC$_o$& 2.49\textperthousand & 90.85 & 96.10 & 68.96 & 90.96 & 94.87 & 88.45 & 89.46 & 91.45 & 88.89 \\  \rowcolor{gray!20}

        \rownumber & AdaLoRA - MPC$_o$& 2.49\textperthousand & 90.59 & 96.10 & 67.87 & 91.16 & 94.67 & 88.45 & 89.71 & 92.08 & 88.83  \\ 
       
         \rownumber & AdaLoRA &  2.49\textperthousand & 90.69 & 96.22 & 67.08 & 91.01 & 94.69 & 89.53 & 90.68 & 92.16 & 89.01 \\  \rowcolor{gray!20}

         \rownumber & FLoRA &  2.49\textperthousand & 90.62 & 96.10 & 68.97 & 91.16 & 94.67 & 87.55 & 89.46 & 92.26 & 88.85 \\ \hline
        
         \rownumber & DoRA & 3.12\textperthousand & 90.60 & 96.22 & 66.62 & 91.23 & 94.47 & 88.08 & 90.93 & 92.00 & 88.77 \\
         \midrule

         \rownumber & HAdapter & 4.57\textperthousand & 90.64 & 96.44 & 67.51 & 91.13 & 94.65 & 88.08 & 90.93 & 92.21 & 88.95 \\ \rowcolor{gray!20}
        
         \rownumber & PAdapter & 4.50\textperthousand & 90.31 & 96.22 & 68.17 & 89.84 & 94.51 & 88.44 & 90.93 & 92.21 & 88.83\\ 

         \rownumber & PA & 4.98\textperthousand & 90.52 & 95.87 & 66.56 & 89.42 & 94.75 & 87.72 & 91.18 & 92.24 & 88.53 \\ \rowcolor{gray!20}

         \rownumber & LoRA & 4.98\textperthousand & 90.68 & 96.10 & 68.36 & 91.16 & 94.60 & 85.20 & 90.93 & 92.43 & 88.68 \\ 
         
         \rownumber & LoRA + MPC$_o$& 4.98\textperthousand & 90.68 & 95.99 & 68.95 & 91.30 & 94.58 & 86.64 & 91.42 & 92.35 & 88.94 \\ \rowcolor{gray!20}

         \rownumber & LoRA + MPC$_d$ & 4.98\textperthousand & 90.72 & 96.21 & 69.55 & 91.20 & 94.47 & 86.28 & 90.44 & 92.27 & 88.89 \\
        
        \rownumber & LoRA + MPC$_n$ & 4.98\textperthousand & 90.52 & 95.87 & 66.56 & 89.42 & 94.75 & 87.72 & 91.18 & 92.24 & 88.53 \\  \rowcolor{gray!20}

        \rownumber & TriLoRA & 4.98\textperthousand & 90.51 & 96.10 & 67.98 & 91.20 & 94.33 & 86.64 & 91.42 & 91.86 & 88.76 \\

        \rownumber & TriLoRA + MPC$_o$& 4.98\textperthousand & 90.96 & 95.76 & 67.26 & 91.08 & 95.06 & 88.81 & 89.95 & 91.80 & 88.84 \\ \rowcolor{gray!20}

        \rownumber & AdaLoRA - MPC$_o$& 4.98\textperthousand & 90.65 & 96.22 & 66.03 & 91.24 & 94.86 & 86.28 & 90.44 & 92.05 & 88.47 \\

        \rownumber & AdaLoRA & 4.98\textperthousand & 90.86 & 95.76 & 65.39 & 90.91 & 94.87 & 89.17 & 89.95 & 91.87 & 88.60 \\  \rowcolor{gray!20}

        \rownumber & FLoRA & 4.99\textperthousand & 90.41 & 95.99 & 69.19 & 91.16 & 94.55 & 87.73 & 89.71 & 92.04 & 88.85 \\ \hline

        \rownumber & DoRA & 5.61\textperthousand & 90.61 & 96.79 & 68.22 & 91.57 & 94.44 & 86.64 & 91.18 & 92.55 & 89.00 \\

        \midrule

         \rownumber & HAdapter & 9.00\textperthousand & 90.57 & 96.10 & 68.57 & 91.62 & 94.73 & 87.00 & 90.93 & 92.49 & 89.00 \\ \rowcolor{gray!20}

         \rownumber & PAdapter & 8.93\textperthousand & 90.29 & 95.76 & 68.47 & 91.39 & 94.60 & 87.72 & 90.93 & 92.61 & 88.97\\

         \rownumber & PA & 9.97\textperthousand & 90.44 & 96.67 & 66.93 & 91.03 & 94.91 & 87.00 & 90.93 & 92.23 & 88.77 \\ \rowcolor{gray!20}

         \rownumber & LoRA & 9.97\textperthousand & 90.73 & 95.87 & 67.83 & 91.28 & 94.73 & 86.28 & 90.69 & 92.30 & 88.71\\
        
         \rownumber & LoRA + MPC$_o$& 9.97\textperthousand & 90.66 & 95.99 & 71.56 & 91.43 & 94.65 & 85.92 & 90.69 & 92.31 & 89.15 \\ \rowcolor{gray!20}

         \rownumber & LoRA + MPC$_d$ & 9.97\textperthousand & 90.64 & 96.22 & 69.23 & 91.22 & 94.91 & 87.00 & 91.67 & 92.24 & 89.14 \\
        
         \rownumber & LoRA + MPC$_n$ & 9.97\textperthousand & 90.44 & 96.67 & 66.93 & 91.03 & 94.91 & 87.00 & 90.93 & 92.23 & 88.77 \\  \rowcolor{gray!20}
         
        \rownumber & TriLoRA & 9.97\textperthousand & 90.41 & 96.67 & 68.01 & 91.46 & 94.58 & 86.64 & 89.95 & 92.35 & 88.76 \\

        \rownumber & TriLoRA + MPC$_o$& 9.97\textperthousand & 90.86 & 96.10 & 66.76 & 91.15 & 95.08 & 88.08 & 90.93 & 91.91 & 88.86 \\ \rowcolor{gray!20}

        \rownumber & AdaLoRA - MPC$_o$& 9.97\textperthousand & 90.83 & 96.22 & 68.36 & 91.25 & 94.82 & 88.45 & 91.42 & 91.92 & 89.16 \\ 

        \rownumber & AdaLoRA & 9.97\textperthousand & 90.65 & 96.10 & 65.96 & 90.80 & 94.73 & 89.53 & 89.71 & 91.90 & 88.67 \\  \rowcolor{gray!20}
        
        \rownumber & FLoRA & 9.98\textperthousand & 90.51 & 95.87 & 67.43 & 91.30 & 94.80 & 88.80 & 90.69 & 92.18 & 88.95 \\ \hline        

         \rownumber & DoRA & 10.59\textperthousand & 90.67 & 95.99 & 66.87 & 91.69 & 94.75 & 87.00 & 91.42 & 92.30 & 88.84 \\

        \midrule

        \rownumber &  HAdapter & 17.87\textperthousand & 90.42 & 96.10 & 67.32 & 91.65 & 94.78 & 89.89 & 89.71 & 92.50 & 89.05 \\ \rowcolor{gray!20}
        
         \rownumber & PAdapter & 17.80\textperthousand & 90.38 & 94.84 & 67.70 & 91.60 & 94.67 & 88.44 & 88.48 & 92.30 & 88.55 \\ 

         \rownumber & PA & 19.94\textperthousand & 90.31 & 96.33 & 66.52 & 91.16 & 94.73 & 86.64 & 91.18 & 92.79 & 88.71\\ \rowcolor{gray!20}

        \rownumber & LoRA & 19.94\textperthousand & 90.69 & 95.87 & 68.36 & 91.35 & 94.62 & 88.08 & 90.69 & 92.05 & 88.96 \\

         \rownumber & LoRA + MPC$_o$& 19.94\textperthousand & 90.77 & 96.10 & 68.74 & 91.44 & 94.97 & 87.73 & 89.95 & 92.27 & 89.00 \\  \rowcolor{gray!20}

         \rownumber & LoRA + MPC$_d$ & 19.94\textperthousand & 90.77 & 96.33 & 68.57 & 91.20 & 94.69 & 87.73 & 90.69 & 92.52 & 89.06 \\
        
        \rownumber & LoRA + MPC$_n$ & 19.94\textperthousand & 90.31 & 96.33 & 66.52 & 91.16 & 94.73 & 86.64 & 91.18 & 92.79 & 88.71 \\ \rowcolor{gray!20}
        
        \rownumber & TriLoRA & 19.94\textperthousand & 90.58 & 95.99 & 69.04 & 91.73 & 94.93 & 88.08 & 90.20 & 92.00 & 89.07 \\ 

        \rownumber & TriLoRA + MPC$_o$& 19.94\textperthousand & 90.83 & 96.44 & 68.57 & 91.27 & 95.02 & 88.81 & 89.95 & 92.05 & 89.12 \\ \rowcolor{gray!20}

        \rownumber & AdaLoRA - MPC$_o$& 19.94\textperthousand & 90.64 & 96.33 & 65.83 & 91.39 & 94.86 & 87.73 & 90.20 & 91.84 & 88.60 \\ 

        \rownumber & AdaLoRA & 19.94\textperthousand & 90.50 & 95.87 & 67.43 & 90.53 & 94.98 & 90.25 & 90.20 & 91.83 & 88.95 \\ \rowcolor{gray!20}

        \rownumber & FLoRA & 19.98\textperthousand & 90.61 & 96.10 & 70.16 & 90.57 & 94.76 & 89.53 & 89.95 & 92.24 & 89.24 \\\hline

        \rownumber & DoRA & 20.56\textperthousand & 90.72 & 96.22 & 68.33 & 91.80 & 94.98 & 88.08 & 90.44 & 92.65 & 89.15\\
         
        \bottomrule
        
    \end{tabular}
    }
    \label{tab: roberta  large results}
\end{table*}

\newpage

\begin{table*}[!ht]
    \centering
    \caption{Details of GLUE dataset.}

    \resizebox{\textwidth}{!}{
    \begin{tabular}{l | l  c  c  c  c  c}
    \toprule
         Dataset & Task & \# Train & \# Dev & \# Test & \# Label & Metrics \\ \midrule
         \multicolumn{7}{c}{Single-Sentence Classification} \\ \hline
         
         CoLA & Acceptability & 8.5k & 1k & 1k & 2 & Matthews corr \\ \midrule
         
         SST-2 & Sentiment & 67k & 872 & 1.8k & 2 & Accuracy \\ \midrule
         
         \multicolumn{7}{c}{Similarity and Paraphrase} \\ \midrule

         MRPC & Paraphrase & 3.7k & 408 & 1.7k & 2 & Accuracy / F1 \\ \midrule

         QQP & Paraphrase & 364k & 40k & 391k & 2 & Accuracy / F1 \\ \midrule
         
         STS-B & Similarity & 7k & 1.5k & 1.4k & 1 & Pearson/ Spearman Corr \\  \midrule

        \multicolumn{7}{c}{Natural Language Inference} \\ \midrule
          
         MNLI & NLI & 393k & 20k & 20k & 3 & Accuracy \\ \midrule
         
         QNLI & QA/NLI & 108k & 5.7k & 5.7k & 2 & Accuracy \\ \midrule

         RTE & NLI & 2.5k & 276 & 3k & 2 & Accuracy \\
        
         \bottomrule
    \end{tabular}}
    \label{tab: glue dataset}
\end{table*}

\begin{table*}[ht]
    \centering
    \caption{Hyper-parameter setup for GLUE benchmark.}
    \resizebox{\textwidth}{!}{
    \begin{tabular}{c|c c c c c c c c }
    \toprule
       Dataset & MNLI & RTE & QNLI & MRPC & QQP & SST-2 & CoLA & STS-B  \\ \hline
       
       learning rate & $5\times10^{-4}$ & $1.2\times10^{-3}$ & $1.2\times10^{-3}$ & $1\times10^{-3}$ &  $5\times10^{-4}$ & $8\times10^{-4}$ & $5\times10^{-4}$ & $2.2\times10^{-3}$\\
       batch size & 32 & 32& 32& 32 & 32 & 32& 32 & 32\\
        \#epochs  & 7 & 50 & 5 & 30 & 5 & 24 & 25 & 25  \\
        
        \bottomrule
    \end{tabular}}
    
    \label{tab:training detail}
\end{table*}


 





\end{document}